\documentclass[letterpaper]{article} 
\usepackage[preprint]{aaai2027}  
\usepackage[hyphens]{url}  
\usepackage{graphicx} 
\usepackage{natbib}  
\usepackage{caption} 
\usepackage{algorithm}
\usepackage{algorithmic}
\usepackage{amsmath}
\usepackage{amssymb}
\usepackage{multirow}
\usepackage{newfloat}
\usepackage{listings}
\DeclareCaptionStyle{ruled}{labelfont=normalfont,labelsep=colon,strut=off} 
\floatstyle{ruled}
\newfloat{listing}{tb}{lst}{}
\floatname{listing}{Listing}

\usepackage{booktabs}

\usepackage{enumitem}
\usepackage{soul}

\nocopyright 

\usepackage{longtable}
\usepackage{booktabs}
\usepackage{multirow}
\usepackage[most]{tcolorbox}
\usepackage{xcolor}
\newcommand{\sout}[1]{  \leavevmode
  \begingroup
  \setbox0=\hbox{#1}  \rlap{\raisebox{0.55ex}{\rule{\wd0}{0.4pt}}}  \box0
  \endgroup
}
\usepackage{multicol}

\newtcolorbox{casebox}[1]{
breakable,
enhanced,
colback=gray!10,
colframe=gray!60,
title=#1,
fonttitle=\bfseries,
coltitle=black,
boxrule=0.8pt,
arc=2mm,
left=2mm,
right=2mm,
top=1mm,
bottom=1mm
}
\usepackage{graphicx}
\title{Learning Simple Test-Time Environments for LLM Web Agents}

\author{
Junxuan Li\textsuperscript{\rm 1*\textdagger} \quad
Zijun Liu\textsuperscript{\rm 2*} \quad
Ziyi Huang\textsuperscript{\rm 3\textdagger} \quad
Peng Li\textsuperscript{\rm 4} \quad
Yuzhou Liu\textsuperscript{\rm 1} \quad
Ming Yan\textsuperscript{\rm 5} \quad
Yang Liu\textsuperscript{\rm 2,4}
}

\affiliations{
\textsuperscript{\rm 1}College of Computer Science and Technology, Jilin University, Jilin, China\\
\textsuperscript{\rm 2}Dept. of Comp. Sci. \& Tech., Tsinghua University, Beijing, China\\
\textsuperscript{\rm 3}School of Electronic and Information Engineering, Beijing Jiaotong University, Beijing, China\\
\textsuperscript{\rm 4}Institute for AI Industry Research (AIR), Tsinghua University, Beijing, China\\
\textsuperscript{\rm 5}Tongyi Lab, Alibaba Group\\
\textsuperscript{*}Equal contribution.
\textsuperscript{\textdagger}Work done at AIR, Tsinghua University.
}

\begin{document}

\maketitle

\begin{abstract}
Large language model (LLM) agents have demonstrated remarkable proficiency in manually constructed environments, yet their performance frequently collapses when transitioned to complex real-world settings. Existing research largely attribute this degradation to the compositional generalization gaps in LLMs on combinations of multiple simple, well-structured environments. In this work, we propose that LLM web agents can learn simple environment observations at test time. Specifically, we introduce trial steps for agents to decompose a complex environment observation into sub-modules, and implement a label-free learning method, Test-Time Environment Decomposition (\textbf{TTED}), to adapt agent behaviors with experience during inference. Our empirical evaluations demonstrate the framework's efficacy across both synthetic and realistic benchmarks, showing (1) experience gains acquired within simpler sub-environments can be effectively composed to improve performance in the full one, and (2) test-time training on sub-environments can significantly enhance the compositional generalization of agents in real-world web automation tasks. 
We also provide key insights in the design of the label-free learning algorithm. 
As more complex environments are accessed by LLM agents, we believe learning environment decomposition skills at test time will be critical for robust real-world deployment.\footnote{Code and data are released at \url{https://github.com/THUNLP-MT/TTED}.} 
\end{abstract}

\begin{figure*}[t]
  \includegraphics[width=\textwidth]{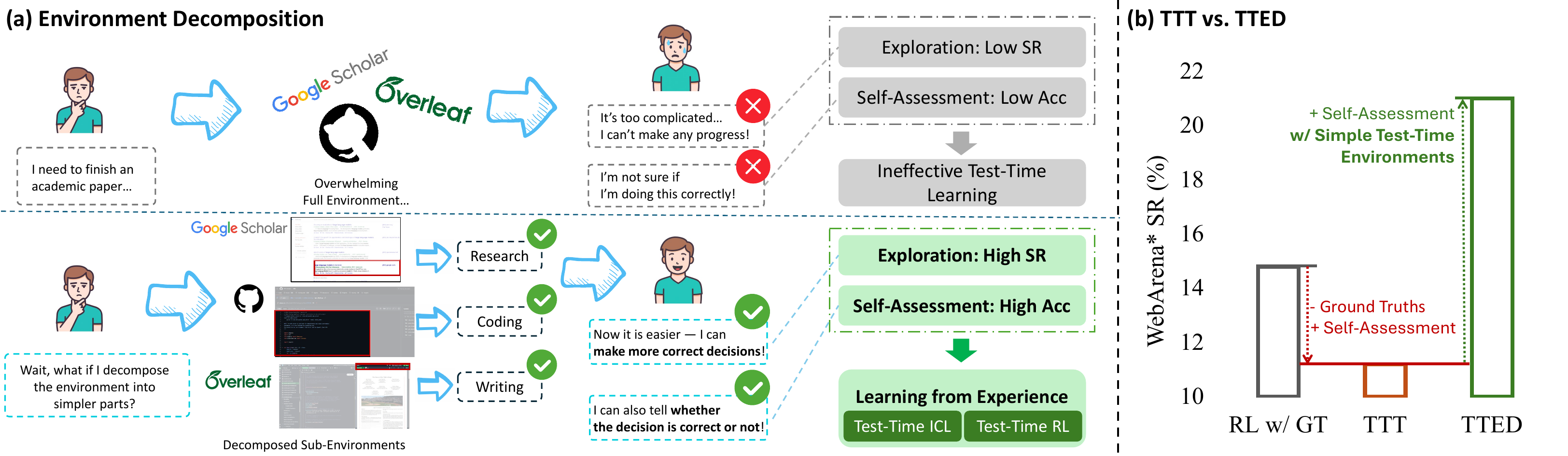}
  \caption{An overview of (a) environment decomposition and a performance comparison of (b) test-time training (TTT) and TTED. ``SR'' denotes success rate. ``GT'' denotes ground truth.  (a) In complex real-world tasks, e.g., writing academic papers, the environment is often compositionally complex. Decomposing into simple test-time environments can help the agent explore and self-assess more effectively, resulting in improved performance after test-time learning. (b) TTED recovers performance degradation of TTT and outperforms RL on full ground-truth rewards. *We only use a subset of WebArena for evaluation.}
  \label{fig:teaser}
\end{figure*}

\section{Introduction}

Large language model (LLM) agents have recently shown strong capabilities in sequential decision making~\citep{Xi2025Rise,OpenAI2026GPT54}, especially in environments where the state space is manually curated, interface layouts are regularized, and task structure is relatively local~\citep{pmlr-v70-shi17a,deng2023mindweb,towers2025gymnasium,liu2025agentenvironmentalignmentautomatedinterface}. Such progress has led to growing optimism that language agents can serve as a general interface for web automation~\citep{he-etal-2024-webvoyager,wei-etal-2025-webagent} and other tasks in diverse interactive environments~\citep{wang2025uitars2technicalreportadvancing,he2025advancing}. 

However, a substantial gap remains between performance in these controlled settings and performance in realistic environments~\citep{zhou2024webarena,pmlr-v235-drouin24a}. 
A central reason for this failure is that real-world environments are compositionally more complex than synthetic ones. In realistic web websites, a single observation may simultaneously contain multiple functional sub-structures, including search panels, form fields, content containers, modal windows, and other task-specific widgets. While each sub-structure may be easy to handle in isolation, their joint presence challenges the compositional generalization ability of current LLM agents. As a result, an agent that succeeds in individual sub-environments may still collapse in a cluttered interface~\citep{furuta2024exposing,boisvert2024workarena}.

Through large-scale post-training~\citep{shen2025thinking,OpenAI2026GPT54}, the base policy can be substantially improved, but further scaling is ultimately constrained by the limited availability of reliable ground-truth reward signals. 
Existing approaches seek to address this problem through more sophisticated test-time learning strategies~\citep{pan2024autonomous,yang2025agentoccam,erdogan2025planandact,chae2025web,he2025advancing,lee2026agentictesttimescalingwebagents}, including prompting, sampling, planning, etc. These methods improve task performance during inference, without external reward feedback. However, the core compositional generalization gap remains. As shown in the example in Figure~\ref{fig:teaser} (a), (1) the agent's exploration success rate degrades when the environment is compositionally complex, resulting in bottlenecked exploration capabilities; and (2) the agent's self-assessment~\citep{zheng2023judging} accuracy also degrades, co-leading to ineffective test-time learning. These limitations become particularly severe in web automation, where relevant decision variables are often localized, while irrelevant context is abundant. However, as an intuitive but unexplored solution, it remains unclear \textbf{whether an agent can learn to simplify environment observations at test time and thus mitigate these failures}.

In this work, we propose \textbf{T}est-\textbf{T}ime \textbf{E}nvironment \textbf{D}ecomposition (\textbf{TTED}), a label-free learning framework that adapts agent behavior within decomposed sub-environments during inference. 
Instead of asking the agent to directly solve the full environment in one pass, we introduce an explicit trial step that the agent partitions a complex observation into a set of specialized and manageable sub-modules. These sub-modules serve as local decision contexts in which the agent can refine its behavior before returning to the full task. Specifically, TTED adapts the agent policy using self-assessment feedback derived from interaction experience within the sub-environments, either by in-context learning (ICL)~\citep{10.5555/3495724.3495883,dong-etal-2024-survey}, which works for single-turn, static webpages, or by reinforcement learning (RL), which collects experience across multiple steps and tasks to fit multi-turn realistic web environments. 

We evaluate TTED on both synthetic and realistic web benchmarks. We extend task quantity in CompWoB~\citep{furuta2024exposing} with an automated synthesis engine to feature more complex environment compositions, resulting in \textit{CompWoB+}. Empirically, we show that TTED with ICL can substantially alleviate the performance degradation when environment complexity increases, showing the capacity of LLMs to perform effective decomposition. We further show that TTED with RL can significantly improve generalization on realistic web automation tasks on WebArena~\citep{zhou2024webarena}, even outperforming RL trained with ground-truth rewards (Figure~\ref{fig:teaser} (b)). We also provide detailed analysis on LLM self-assessment quality, out-of-distribution generalization on WorkArena~\citep{pmlr-v235-drouin24a}, data and environment scaling effects, and the design of RL algorithms.

In summary, our key contributions are as follows:
\begin{itemize}
    \item We introduce TTED as a test-time learning solution to compositional generalization failures in LLM web agents. 
    \item We extend a larger synthetic benchmark, CompWoB+, to systematically evaluate the impact of environment composition on agent performance and show that TTED can effectively mitigate the resulting generalization gap.
    \item We demonstrate TTED with proper RL design is effective in realistic web automation tasks such as WebArena, even outperforming RL trained with ground-truth rewards.
    \item We analyze several key factors that drive the success of test-time adaptation under environment decomposition.
\end{itemize}
\section{Related Work}
\label{sec:related_work}

\paragraph{Environment Generalization in LLM Web Agents}
The transition from synthetic to real-world environments exposes a critical vulnerability in LLM agents: environment generalization.
LLMs can recombine learned primitives in controlled settings~\citep{wei2022chain,abedsoltan2025task}, and multi-agent RL can improve transfer across distinct environments~\citep{he2025advancing}.
However, recent work~\citep{boisvert2024workarena,xu-etal-2025-crab} suggests that LLMs often exhibit brittle generalization when tasks require systematic composition across multiple web modules, pages, or websites.
This challenge is acute in realistic websites, where task-relevant modules are buried in irrelevant context; even strong proprietary models remain weak on end-to-end knowledge-work automation~\citep{boisvert2024workarena}.
Maintaining both reliable action generation and accurate self-assessment in such cluttered environments therefore remains an open challenge.

\paragraph{Test-Time Learning for LLM Web Agents}
\label{sec2:TTL}
Test-time learning adapts agent behavior during inference using signals available from test instances rather than ground-truth labels~\citep{pmlr-v267-akyurek25a,yuksekgonul2026learningdiscovertesttime}.
For web agents, these methods include prompting, sampling, planning, and reflection~\citep{pan2024autonomous,erdogan2025planandact,chae2025web,lee2026agentictesttimescalingwebagents}.
ICL~\citep{10.5555/3495724.3495883} methods optimize the agent's behavior at test time on each query individually, which can be termed as \emph{query-level} test-time learning. However, they are constrained to benefit from experience within multiple different queries. 
Test-time training~\citep{pmlr-v267-akyurek25a,zuo2025ttrl} instead updates model parameters from experience accumulated across queries.
However, existing approaches either assume synthetic task formats~\citep{pmlr-v267-akyurek25a} or rely on matching-based self-consistency rewards~\citep{zuo2025ttrl}, which cannot directly supervise free-form environment decomposition.
Although test-time training can improve generalization under distribution shift~\citep{pmlr-v119-sun20b}, its effectiveness depends on accurate and informative learning signals, which are often sparse and noisy in complex environments.

\paragraph{Positioning of Our Work}
TTED connects these directions by decomposing a complex environment into an aligned sub-goal and sub-observation, collecting self-assessed experience in this local context, and adapting the agent through query-level ICL or task-level RL.
Unlike planning and reflection~\citep{erdogan2025planandact,pan2024autonomous} typically modify reasoning, action search, or feedback while retaining the full environment representation, TTED decomposes the representation for both action generation and self-assessment.
It also differs from history shrinking or context pruning~\citep{yang2025agentoccam}: localized sub-environments are not merely shorter inputs but test-time learning units.
TTED therefore directly targets the coupled degradation of exploration and self-assessment under compositional complexity.
Further distinctions from planning are discussed in Appendix~A.1.

\section{Preliminaries}
\label{sec:preliminaries}

\begin{figure}[t]
  \centering
  \includegraphics[width=0.47\linewidth]{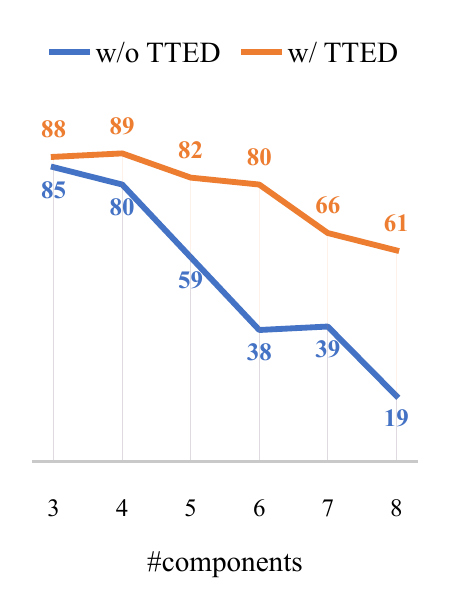}
  \hfill
  \includegraphics[width=0.47\linewidth]{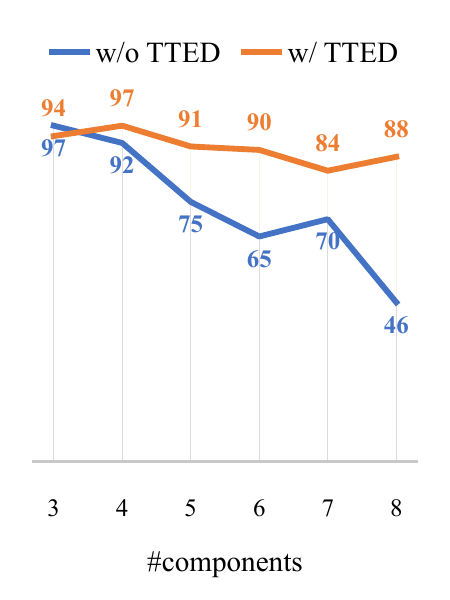}
  \caption{Task success rates (\%) on CompWoB+ with Qwen3-8B (left) and GPT-4o-mini (right). TTED effectively mitigates compositional generalization failures on both models.}
  \label{fig:compwob}
\end{figure}

\subsection{Test-Time Environment}

We model the test-time environment of a long-horizon web task as a Markov decision process (MDP) $\mathcal{E}=(\mathcal{S},\mathcal{A},\mathcal{T},\mathcal{R})$. A test-time environment refers to the task-conditioned decision process encountered by the policy during inference, including the interaction context observable to the agent, the available actions, the environment transitions, and the reward signals associated with the task. At step $t$, the state is
\begin{equation}
\label{eq:state}
s_t=(g,o_t,H_{t-1}),
\end{equation}
where $g$, $o_t$, and $H_{t-1}$ denote the task goal, current observation, and interaction history, respectively. A policy $\pi_\theta(a_t\mid s_t)$ induces a trajectory $\tau=(o_1,a_1,\ldots,o_T,a_T)$. A conventional approach directly optimizes the policy in the original test-time environment:
\begin{equation}
\pi_{\mathrm{ori}}^*
=
\arg\max_{\pi}
\mathbb{E}_{\tau\sim\pi}
[\mathcal{R}(\tau)].
\end{equation}
However, abundant irrelevant information and multi-step execution in complex web environments make task rewards sparse and delayed, while also complicating accurate step-level reward estimation. Consequently, direct policy optimization in the full test-time environment is difficult.

To address this issue, TTED decomposes the original test-time environment into multiple aligned local decision contexts, each consisting of a sub-goal and its task-relevant sub-observation. This process yields a set of localized test-time environments $\{\mathcal{E}^{(i)}\}_{i=1}^{n}$, where
\begin{equation}
\mathcal{E}^{(i)}
=
(\mathcal{S}^{(i)},\mathcal{A},\mathcal{T},\mathcal{R}^{(i)}).
\end{equation}
Each localized test-time environment restricts the state and decision context while preserving the original action space and transition dynamics. Since it contains less irrelevant information and lower decision ambiguity, its local reward signal can be estimated more reliably. TTED therefore optimizes
\begin{equation}
\pi_{\mathrm{TTED}}^*
=
\arg\max_{\pi}
\sum_{i=1}^{n}
\mathbb{E}_{\tau^{(i)}\sim\pi}
[\mathcal{R}^{(i)}(\tau^{(i)})].
\end{equation}
By accumulating learning gains across these localized test-time environments, TTED replaces difficult direct optimization in the original complex environment with more tractable local optimization. The resulting local policy improvements are subsequently composed to solve the full task.

\begin{figure*}[t]
  \centering
  \includegraphics[width=0.9\linewidth]{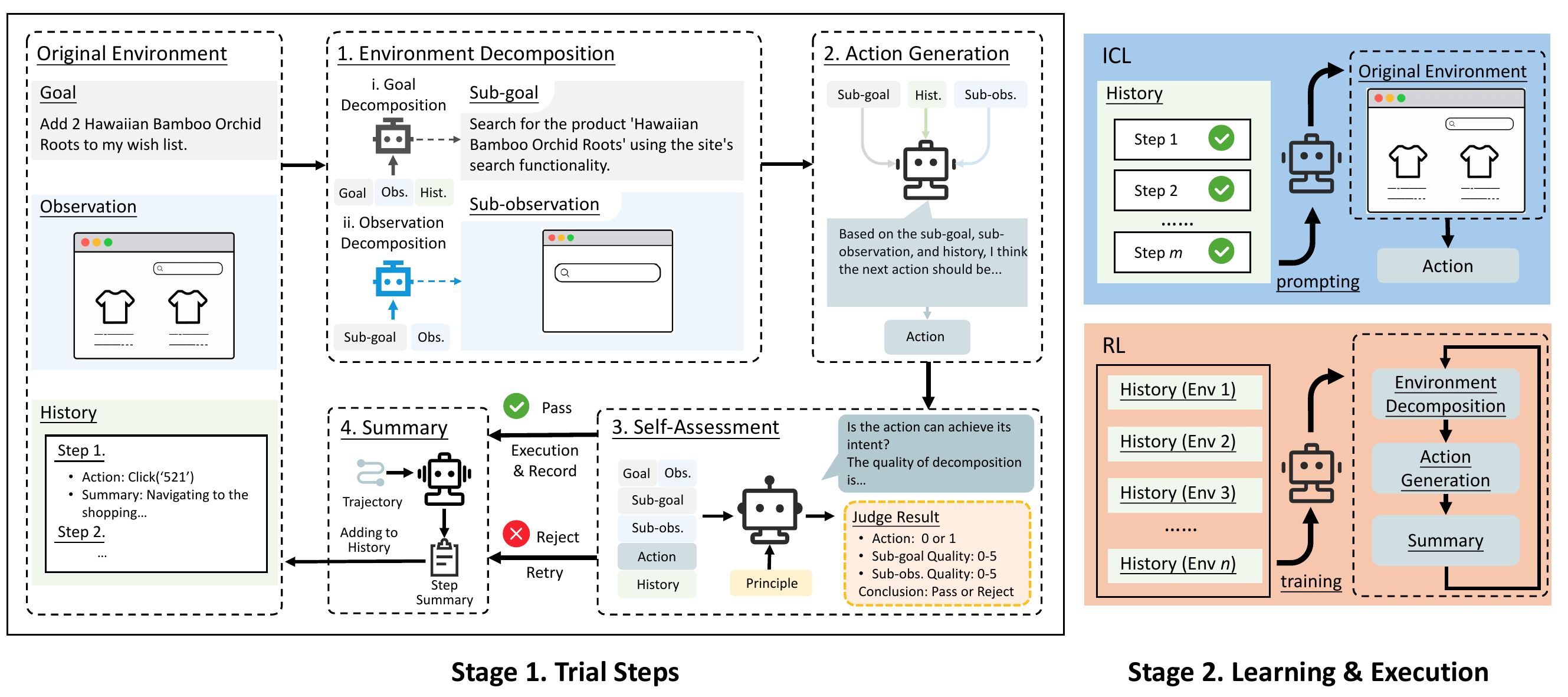}
  \caption{The Test-Time Environment Decomposition (TTED) methodology, illustrating the two-stage process of (1) Trial Steps with environment decomposition and (2) Test-Time Learning and Execution, utilizing either ICL or RL. The agent gathers experience within the decomposed sub-environments through trials. And ICL leverages correct past actions as task-specific demonstrations to solve tasks within single-turn, static webpages, while RL optimizes both decomposition process and action generation policy with trajectories from multiple environments, enabling adaptation in multi-turn, realistic web environments.}
  \label{fig:method}

\end{figure*}

\subsection{Test-Time Learning for LLMs}
\label{sec:ttl}

Test-time learning updates a model during inference using signals available from test instances, without ground-truth labels.
For LLM agents, such signals may come from self-consistency~\citep{zuo2025ttrl}, self-assessment~\citep{zheng2023judging}, or external feedback~\citep{pan2024autonomous,chae2025web}.
Given a pretrained policy $\pi_{\theta_0}$ and test instances $\mathcal{X}$, adaptation is written as
\begin{equation}\label{eq:ttl-general}
\pi_{\theta'}
\leftarrow
\mathcal{F}(\pi_{\theta_0};\mathcal{X}),
\end{equation}
where $\mathcal{F}$ uses unlabeled or weakly supervised inference-time signals.
TTED instantiates $\mathcal{F}$ through either ICL or RL.

\paragraph{In-Context Learning (ICL)}
ICL~\citep{10.5555/3495724.3495883,dong-etal-2024-survey} adapts behavior without updating $\theta_0$. TTED collects successful state--action pairs during trial steps as a demonstration set $\mathcal{C}=\{(s_j,a_j)\}_{j=1}^{m}$, and conditions subsequent action generation on this self-generated experience:
\begin{equation}\label{eq:icl}
\pi_{\mathrm{ICL}}(a_t\mid s_t)
=
\pi_{\theta_0}(a_t\mid s_t,C).
\end{equation}
This query-level adaptation reuses verified local successes when acting in the original complex environment.

\paragraph{Reinforcement Learning (RL)}
For multi-turn web tasks, TTED collects trajectories across interactions and updates the model parameters.
Because ground-truth step rewards are unavailable at test time, $r_t$ is replaced by an assessed reward $r_t^*$.
The reward can usually be derived by matching-based self-consistency~\citep{zuo2025ttrl} or generative reward models~\citep{zheng2023judging,liu2025inferencetimescalinggeneralistreward}.
Using undiscounted REINFORCE~\citep{williams1992simple}, we estimate
\begin{equation}\label{eq:reinforce}
\nabla_\theta J(\theta)
=
\mathbb{E}_{\tau\sim\pi_\theta}
\left[
\sum_{t=1}^{T}
\nabla_\theta
\log\pi_\theta(a_t\mid s_t)
r_t^*
\right].
\end{equation}
The adapted parameters are then used for final execution in the full environment.
The following methodology section specifies how TTED constructs $r_t^*$ from environment decomposition and action assessment.

\subsection{Compositional Generalization Gaps in LLMs}
\label{sec:compwob}

Compositional generalization asks whether a model can solve novel combinations of familiar components, including interface motifs, interaction primitives, layouts, and instruction patterns.
To examine this gap in web agents, we extend CompWoB~\citep{furuta2024exposing}, which composes ten base web modules but contains few tasks exceeding five components.
Our extension, \textit{CompWoB+}, uses an automated synthesis engine while retaining the original reward rules.
We evaluate Qwen3-8B~\citep{yang2025qwen3technicalreport} and GPT-4o-mini~\citep{openai2024gpt4ocard} on 100 tasks per composition size from 3 to 8, yielding 600 tasks in total.
Further details appear in Appendix~D.

As shown in Figure~\ref{fig:compwob}, increasing the number of components from 3 to 8 reduces the success rate of Qwen3-8B from 85\% to 19\% and that of GPT-4o-mini from 97\% to 46\%.
Because these tasks recombine familiar base modules under the original reward rules, the decline is consistent with a compositional generalization gap when components co-occur in a single observation.
With TTED, the corresponding success rates at eight components increase to 61\% and 88\%, motivating the framework introduced next.

\section{TTED: Test-Time Environment Decomposition}

\begin{figure*}[t]
  \centering
\begin{minipage}{0.59\textwidth}
  \centering
  \resizebox{\linewidth}{!}{
  \begin{tabular}{llrcccccc}
    \toprule
    \multicolumn{2}{l}{\multirow{2}{*}{\textbf{Method}}} & \textbf{\#Sub-} & \multicolumn{5}{c}{\textbf{WebArena}} & \multirow{2}{*}{\textbf{WorkArena}}\\ \cmidrule{4-8}
    & & \textbf{Envs} & CMS &  Map & Shopping & Gitlab & \textbf{Overall} & \\
    \midrule
    \multicolumn{2}{l}{\textcolor{gray}{Qwen3-8B}} & \textcolor{gray}{0} & \textcolor{gray}{12.1} & \textcolor{gray}{8.1} & \textcolor{gray}{14.9} & \textcolor{gray}{12.3} & \textcolor{gray}{12.1} & \textcolor{gray}{18.2} \\ \midrule
    \multicolumn{2}{l}{SFT w/ GT} & 3200 & 11.4 & 12.1 & 17.8 & 14.0 & 14.1 & 17.9 \\ 
    \multicolumn{2}{l}{RL w/ GT} & 3200 & 12.1 & 15.3 & 17.8 & 13.2 & 14.8 & 17.3 \\ 
    \multicolumn{2}{l}{TTRL} & 3200 & 11.4 & 11.3 & 15.5 & 15.8 & 13.5 & - \\ 
    \multirow{2}{*}{TTT} & SFT  & 3200 & 10.7 & 9.7 & 12.1 & 12.3 & 11.2 & 18.2  \\ 
     & RL & 3200 & 10.7 & 10.5 & 13.8 & 14.9 & 12.5 & 17.3 \\ 
    \midrule
    \multirow{3}{*}{\textbf{TTED}} & SFT & 3200 & 17.4 & 12.1 & 20.7 & 19.3 & 17.6 & 21.8 \\
     & RL & 800 & 19.5& 15.3 & 21.8 & 20.2 & 19.4 & - \\ 
     & RL & 3200 & 22.8 & 12.9 & 24.7 & 21.9 & \textbf{21.0} & \textbf{22.7} \\ 
    \bottomrule
  \end{tabular}
  }
  \captionof{table}{Experimental Results for Test-Time Training on WebArena. Success rates (\%) are reported for WebArena and WorkArena (out-of-distribution). ``\#Sub-Envs'' denotes the number of sub-environments the agent has explored on during test-time learning.}
  \label{tab:main-ood}
\end{minipage}
\hfill
\begin{minipage}{0.39\textwidth}
  \centering
  \includegraphics[width=0.7\linewidth]{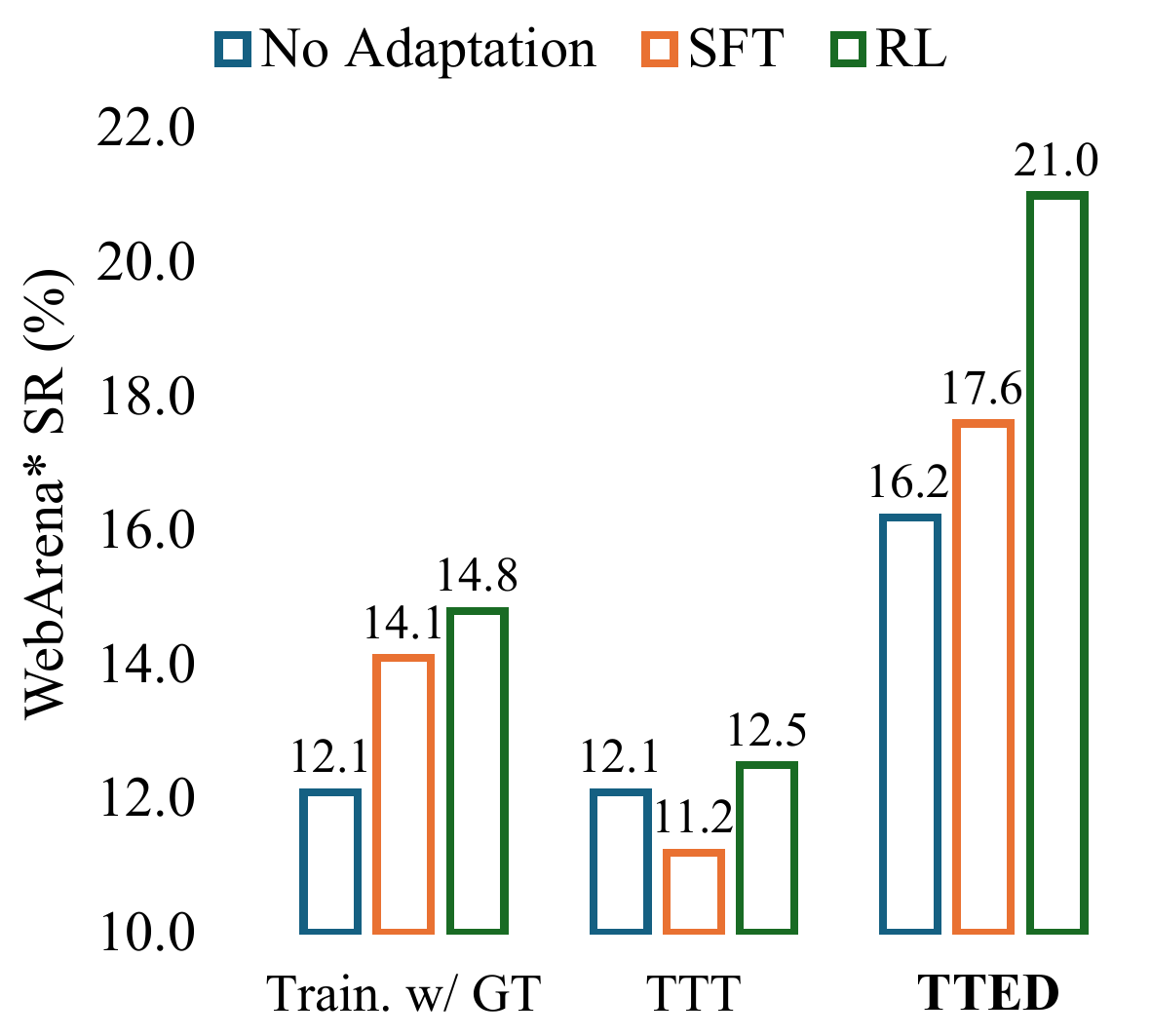}
  \caption{Task success rates (\%) of test-time learning methods on WebArena (*subset). ``No Adaptation'' denotes the base model within the same inference pass.}
  \label{fig:main-no}
\end{minipage}

\end{figure*}

As shown in Figure~\ref{fig:method}, TTED first gathers self-assessed experience through trial steps in decomposed sub-environments, and then adapts the agent through query-level ICL or task-level RL before final execution in the original environment. All prompt templates are provided in Appendix~E.

\subsection{Experience Gathering via Trial Steps}\label{sec:trials}

To maintain a coherent reasoning trace throughout the entire trial process, TTED uses the same LLM across the four trial steps, adopts a higher sampling temperature for diverse exploration, and represents web content as an unfolded accessibility tree following~\citet{yang2025agentoccam}.

\paragraph{Step 1: Environment Decomposition} In the first step, the agent performs decomposition to the environment state $s_t$ (Equation~\eqref{eq:state}) to obtain a sub-environment along with an aligned goal structure. 
Given $s_t=(g,o_t,H_{t-1})$, TTED first predicts a localized sub-goal
\begin{displaymath}
g_t^{(i)}=\pi_{\mathrm{goal}}(g,o_t,H_{t-1}),
\end{displaymath}
and then extracts its task-relevant sub-observation
\begin{displaymath}
o_t^{(i)}=\pi_{\mathrm{obs}}(o_t,g_t^{(i)}).
\end{displaymath}
Together, they define the sub-environment state
\begin{displaymath}
S^{(i)}=(g_t^{(i)},o_t^{(i)},H_{t-1}),
\end{displaymath}
while preserving the original action space and transition dynamics. For interactive websites, this sub-environment is a textual view of the original environment; for static webpages, it is instantiated in an isolated simulator.

\paragraph{Step 2: Action Generation}
Once the sub-environment is constructed, the action agent predicts the next action based on the sub-environment state. Formally, the action generation policy $\pi_{\text{action}}$ produces an action tailored to the sub-goal: 
\begin{displaymath}
a_t^{(i)}=\pi_{\text{action}}(g_t^{(i)}, o_t^{(i)}, H_{t-1}). 
\end{displaymath}

\paragraph{Step 3: Self-Assessment}  
The agent evaluates both the quality of the environment decomposition and the correctness of the generated action as a generative reward model, following a predefined set of principles~\citep{liu2025inferencetimescalinggeneralistreward}. 
The agent assigns scores to the sub-goal and sub-observation, yielding $r_{\text{goal}},r_{\text{obs}}\in[0,5]$, and determines the action reward with a binary score $r_{\text{action}}\in\{0,1\}$. 
Formally:  
\begin{displaymath}
(r_{\text{goal}}, r_{\text{obs}}, r_{\text{action}})
= \mathcal{D}_{\text{judge}}\big(o_t, g_t^{(i)}, o_t^{(i)}, a_t^{(i)},H_{t-1},g\big) . 
\end{displaymath}

If the decomposition scores exceed a predefined threshold and $r_{\text{action}}=1$, the trial step is marked as ``Pass'' and the experience (trajectory $\tau$) is recorded. Otherwise, it is marked ``Reject'' and the step is retried up to a fixed limit.
The self-assessment scores are used for the subsequent learning stage: 
\begin{displaymath}
r_t
=
\left(
r_{\text{goal}},
r_{\text{obs}},
r_{\text{action}}
\right)
\end{displaymath}
e.g., in Equation~\eqref{eq:icl} and Equation~\eqref{eq:reinforce}. 

\paragraph{Step 4: Summary}  
If the trial passes, the agent synthesizes a step summary to serve as a natural-language description of the executed action and its intent: $h_t = \mathcal{D}_{\text{summary}}\bigl(a_t^{(i)}\bigr)$, 
where $\mathcal{D}_{\text{summary}}$ denotes the summary-generation function, which will not be optimized in the learning stage of TTED. 
This updated history $H_t=H_{t-1}\cup\{h_t\}$ manages modest context lengths, informing subsequent steps. 

The environment then transitions according to $s_{t+1} = \mathcal{T}\!\left(s_t,\,a_t^{\text{exec}}\right)$, 
where the executed action $a_t^{\text{exec}}=a_t^{(i)}$ if receiving pass signal from self-assessment, and $a_t^{\text{exec}}=\text{None}$ otherwise, indicating that no action is applied to the environment.
The whole process repeats until a terminal condition is reached or the interaction step limit is exceeded.

\subsection{Test-Time Learning from Experience}\label{sec:learning}

\subsubsection{Specific Designs for Test-Time ICL}

\paragraph{Self-Assessment with Executable Scripts}
For single-turn static webpages, pure LLM self-assessment may produce false positives. 
To enhance reliability, the self-assessment step leverages executable code scripts (e.g., via Selenium) under the decomposed sub-environment to execute and verify the generated action in an isolated simulation, which provides a highly accurate deterministic reward signal $r_{\text{action}}$ for ICL without causing any side effects.

\paragraph{ICL Demonstration Curation}
To ensure the quality of the test-time adaptation on the given task query, TTED strictly curates the execution history from the same task. Only trial steps that are explicitly passed by self-assessment (i.e., $r_{\text{action}} = 1$) are retained as successful demonstrations $C=\{(s_j,a_j)\}_{j=1}^m$. Failed trial steps are discarded. The policy is then conditionally updated to generate final actions based on this high-quality augmented context according to Equation~\eqref{eq:icl}. The full process is illustrated in Appendix~D.

\subsubsection{Specific Designs for Test-Time RL}

\paragraph{Self-Assessment for Multi-turn Execution} 
In realistic multi-turn websites, faithfully replicating and simulating the sub-environment is infeasible, while an incorrect action may irreversibly derail the task. 
Meanwhile, using LLM as a generative reward model allows TTED to flexibly assess the quality of both the decomposition and the action, where the matching-based self-consistency reward design in TTRL~\citep{zuo2025ttrl} only applies to label actions.
After test-time RL, the agent re-executes the same decomposition and action-generation pipeline without self-assessment or retry, which allows the agent to directly apply its learned decomposition and action generation strategies.

\paragraph{Learning from Rejected Trials} 

Unlike ICL, RL retains both passed and rejected trials. This improves data efficiency when correct trajectories are sparse; moreover, because $r_{\mathrm{action}}=0$ need not imply incorrect goal or observation decomposition, the decoupled rewards allow TTED to learn useful decomposition behavior from rejected actions.

\paragraph{Policy Optimization Algorithm Design} 
To optimize the unified model parameters $\theta$ governing both the decomposition process ($\pi_{\text{goal}}, \pi_{\text{obs}}$) and the action generation policy ($\pi_{\text{action}}$), we design an off-policy training scheme. Because the rollouts are generated by an older behavioral policy in the inference engine $\pi_{\theta_{\text{old,inf}}}$ during the trial steps, we utilize rollout importance sampling~\citep{liu-li-2025-rl-collapse} with the ratio $\rho_t=\frac{\pi_\theta(a_t|s_t)}{\pi_{\theta_{\text{old,inf}}}(a_t|s_t)}$. To construct meaningful (avoiding zeros) and less variant gradients, the raw assessment scores are normalized into a bounded return $r^*_{t}\in[-1,1]$. To stabilize the highly variable test-time training process across diverse web layouts, we adopt the advantage normalization technique from REINFORCE++~\citep{hu2025reinforcestabilizingcriticfreepolicy}. The advantage $\hat{A}_t$ is calculated by standardizing the cumulative returns across the training mini batch of collected trial environments (not necessarily from a group of trajectories from the same environment): 
$ \hat{A}_t=\frac{r^*_{t}-\mu_{r^*}}{\sigma_{r^*}+\epsilon} $, 
where $\mu_{r^*}$ and $\sigma_{r^*}$ are the mean and standard deviation of the normalized returns, and $\epsilon$ is a small constant for numerical stability. Inheriting from Equation~\eqref{eq:reinforce}, the final surrogate objective to be maximized is therefore formulated as: 
\begin{equation}
\begin{aligned}
\nabla_\theta J(\theta)
&=
\mathbb{E}_{\tau \sim \pi_{\theta_{\text{old,inf}}}}
\left[
\sum_{t,i}
\sum_{a^*\in\{g_t^{(i)}, o_t^{(i)}, a_t^{(i)}\}}
\right.\\
&\qquad\left.
\rho_t
\nabla_\theta\log\pi_\theta(a^*|s_t)
\hat{A}_t
\right],
\end{aligned}
\end{equation}
where environment decomposition and action generation are all counted in the same interaction step. 
We also empirically observe that this normalization is key to stabilizing advantage estimation and improving off-policy RL scalability.

\section{Experiments and Analysis}\label{sec:exp}

We perform several experiments to evaluate the effectiveness of TTED and to answer a few critical research questions: 
\textbf{RQ1:} \emph{Can TTED mitigate compositional generalization failures beyond synthetic web environments?} 
\textbf{RQ2:} \emph{Which decomposition components are essential for effective label-free test-time learning?} 
\textbf{RQ3:} \emph{What is the key factor that affects the scalability and robustness of TTED?} 
\subsection{Experiment Settings}\label{sec:exp-setup}

\subsubsection{Benchmarks}

We evaluate query-level ICL on \textit{CompWoB+} and task-level RL on \textit{WebArena}~\citep{zhou2024webarena} and \textit{WorkArena}~\citep{pmlr-v235-drouin24a,boisvert2024workarena}.
CompWoB+ provides controlled compositions of web components, whereas WebArena and WorkArena contain heterogeneous, real-world interfaces.
RL adapts once on WebArena and is evaluated both on the same benchmark distribution and on WorkArena without further training.
We use raw DOM observations for CompWoB+ and accessibility-tree observations through BrowserGym for the realistic benchmarks.

\subsubsection{Baselines and Evaluation Protocol}
To evaluate TTED as a \emph{label-free, transductive test-time adaptation} method, not general post-training, we experiment consistently with prior test-time learning methods~\citep{pmlr-v267-akyurek25a,zuo2025ttrl}. 
On realistic web automation benchmarks, 
we compare TTED to a range of test-time learning baselines that do not explicitly decompose the environment: 

\begin{itemize}[nosep]
\item \textbf{No Adaptation}: The base LLM is directly applied to the full environment without any test-time adaptation. 
\item \textbf{Training w/ GT}: LLMs are trained with ground-truth labels from embedded rules on the full environment. 
\item \textbf{Test-Time Training} (TTT)~\citep{pmlr-v267-akyurek25a}: It represents a method similar to TTED but without environment decomposition. 
\item \textbf{Test-Time RL} (TTRL)~\citep{zuo2025ttrl}: TTT method that applies self-consistency reward labeling and GRPO algorithm~\citep{shao2024deepseekmathpushinglimitsmathematical}. 
\end{itemize}

Orthogonal to these baselines, we also include various advantage estimation designs, including Supervised Fine-Tuning (SFT), REINFORCE (without advantage normalization), and GRPO (required by TTRL), to ablate the impact of policy optimization algorithm. 

Evaluation is performed using task success rates. We also record efficiency measures such as numbers of interaction steps, token consumption, and training time. 
For test-time training methods, we control the training budget to ensure fair comparison. 
Since \textbf{constructing true local labels at step for realistic websites is non-trivial}, 
we also include human evaluation to examine the adaptation process of TTED, such as the quality of inferred sub-environments and the consistency between local action decisions and global task goal. 

\subsubsection{Implementation Details}

On CompWoB+, we evaluate Qwen3-8B~\citep{yang2025qwen3technicalreport} with reasoning CoT and GPT-4o-mini~\citep{openai2024gpt4ocard}. On WebArena and WorkArena, we evaluate the open-source Qwen3-8B for model training. Full settings are provided in Appendix~D.

\begin{table}[t]
\resizebox{\linewidth}{!}{
\begin{tabular}{llccccc}
\toprule
\multirow{2}{*}{\textbf{Method}} & \multirow{2}{*}{\textbf{Reward Labeling}} & \multirow{2}{*}{\textbf{\#Steps}} & \multicolumn{4}{c}{\textbf{Reward Consistency}} \\
\cmidrule{4-7}
& & & $r_{\text{goal}}$ & $r_{\text{obs}}$ & $r_{\text{action}}$ & Overall \\
\midrule
TTRL & self-consistency & 4.05 & - & - & 0.68 & 0.68 \\
TTT & principled GRM & 7.18 & - & - & 0.71 & 0.71 \\
\textbf{TTED} & principled GRM & 8.23 & 0.81 & 0.87 & \textbf{0.86} & \textbf{0.85} \\
+ TTRL (action) & self-consistency & 8.44 & - & - & 0.80 & 0.80 \\
\bottomrule
\end{tabular}
}
\caption{Consistency evaluation (normalized AD) of reward labeling before model training from test-time learning methods with humans. Evaluation is on 82 tasks from WebArena.}
\label{tab:human-eval}

\end{table}

\begin{table}[t]
  \centering
  \resizebox{\linewidth}{!}{
  \begin{tabular}{lcccc}
    \toprule
    \textbf{Advantage Estimation} & \textbf{\#Sub-Envs} & Train. w/ GT & TTT & TTED \\
    \midrule
    SFT & 3200 & 14.1 & 11.2 & 17.6 \\ \midrule
    \multirow{4}{2cm}{REINFORCE w/ Advantage Normalization} & 800 & - & 12.5 & 19.4 \\
     & 1600 & - & 13.5 & 18.4 \\
     & 2400 & - & 13.2 & 20.0 \\
     & 3200 & 14.8 & 12.5 & \textbf{21.0} \\
     \midrule
    REINFORCE & 3200 & 13.0 & 14.3 & 17.6 \\
    GRPO & 3200 & - & 13.5 & 17.8 \\
  \bottomrule
\end{tabular}
  }
  \caption{Test-time scaling experiments for test-time learning methods on WebArena with different advantage estimation strategies. We report task success rates (\%) under different training budgets, i.e. numbers of explored sub-environments.}
  \label{tab:scaling-time}
  
\end{table}

\subsection{Main Results}

\paragraph{Mitigating Synthetic Compositional Generalization Failures}
Figure~\ref{fig:compwob} illustrates performance on the synthetic CompWoB+ benchmark as the number of concurrent environment components scales from 3 to 8. As stated in the Preliminaries section, without adaptation, both base models suffer a severe performance collapse as interface complexity increases. However, applying TTED effectively arrests this degradation. After TTED with ICL, Qwen3-8B maintains 80\%+ success rates for less than 6 components, achieving 61\% at 8 components, while GPT-4o-mini retains a high score of 88\%. \textbf{Takeaway:} \ul{The performance drop in cluttered environments is primarily a failure of compositional generalization rather than lacking foundational knowledge, and test-time environment decomposition provides high-quality demonstrations for ICL, which successfully preserve agent capabilities as structural complexity scales.}

\paragraph{Performance on Realistic Web Environments}
As shown in Table~\ref{tab:main-ood} and Figure~\ref{fig:main-no}, TTED significantly outperforms all baselines on WebArena. Without adaptation, the base Qwen3-8B achieves only a 12.1\% success rate. Standard post-training with ground-truth (GT) labels yields marginal gains (14.8\% for RL) potentially because of reward sparsity, while monolithic test-time training methods (TTT, TTRL) fail to surpass these GT baselines due to noisy reward labeling in complex interfaces. Conversely, TTED (RL) achieves superior performance with 1/4 of the training budget, and further scales up to a state-of-the-art 21.0\% success rate with full budget. \textbf{Takeaway:} \ul{Label-free adaptation via decomposed simple test-time environments effectively surpasses monolithic methods both in SFT and RL, even surpassing full ground-truth supervision.} 

\paragraph{Out-of-Distribution Generalization}
To assess cross environment transferability, we evaluated the adapted models on WorkArena without further  training (Table~\ref{tab:main-ood}). TTED (RL) maintains superior performance, achieving a 22.7\% success rate compared to the base model and standard TTT. \textbf{Takeaway:} \ul{The decomposition and interaction policies learned by TTED represent robust, generalizable capabilities.} 

\textbf{Additional model evaluations, significance tests, and efficiency analyses are provided in Appendix~B.}

\begin{table}[t]
  \centering
  \resizebox{0.8\linewidth}{!}{
  \begin{tabular}{lccccc}
    \toprule
    \textbf{Method}&\textbf{WebArena}\\
    \midrule
    \textbf{TTED} &  \textbf{21.0} \\
     - Sub-goal decomposition &  12.4 \\
     - Sub-observation decomposition & 15.0 \\
     - Both = TTT & 12.5 \\
  \bottomrule
\end{tabular}
  }
  \caption{Ablation of TTED components on WebArena, by removing sub-goal and sub-observation decomposition. We report task success rates (\%) under the same RL budget.}
  \label{tab:abl}
  
\end{table}

\subsection{Ablation Studies}
\label{sec:ablation}
\paragraph{Importance of Decomposition Components}
Table~\ref{tab:abl} isolates the individual contributions of the environment decomposition step in TTED. Removing sub-goal decomposition degrades the success rate severely from 21.0\% to 12.4\%, indicating agents disoriented without localized objectives. Removing sub-observation partitioning drops performance to 15.0\%. Removing both reduces the framework to standard TTT (12.5\%). \textbf{Takeaway:} \ul{The joint application of structural and goal decomposition primarily drives TTED's efficacy.} 

\paragraph{Self-Assessment Quality}
Reliable reward signals are a known bottleneck in label-free test-time learning. TTED optimizes both decomposition and action policies. Without LLM-as-a-judge, e.g., TTRL, reward signals cannot cover the trajectory. 
Table~\ref{tab:human-eval} evaluates human agreement with reward labeling prior to model training. For human evaluation, two graduate-level annotators follow guidelines (see Appendix~E) and reaches 0.85 Cohen's $\kappa$. Standard TTRL (self-consistency) and TTT (principled GRM) achieve moderate overall agreements of 0.68 and 0.71, respectively. By contrast, standard TTED reaches an exceptional 0.85. Self-consistency labeling also demonstrates improved quality when performed in decomposed environments (TTED + TTRL (action)). \textbf{Takeaway:} \ul{LLM self-assess as a GRM with proper principles yields high-fidelity signals across goal alignment, observation relevance, and action correctness, which is \textbf{critical} for scalable test-time learning.} 

\paragraph{RL Algorithm Design and Test-Time Scaling}
Table~\ref{tab:scaling-time} investigates test-time compute scaling and policy optimization designs. As the sub-environment budget scales from 800 to 3200, standard TTT plateaus at ~13.5\% due to adaptation drift from noisy global updates. In contrast, TTED scales consistently up to 21.0\%. Crucially, advantage normalization in sample batches stabilizes training, compared to standard REINFORCE or GRPO with peak performance to ~17.8\%. \textbf{Takeaway:} \ul{Cross-environment advantage normalization is essential for stabilizing and scaling off-policy test-time RL.} 

\subsection{Qualitative Observations and Failure Cases}
\label{sec:qual}

We also perform qualitative case analysis to understand how TTED succeeds or fails in different scenarios (see Appendix~C for detail). In summary, qualitative analyses show that environment decomposition mitigates interface clutter by isolating functional components and shifting agents from brute-force interaction toward appropriate tool use. Test-time training further improves robustness to layout variations. Remaining failures usually stem from action-space misalignment or overthinking rather than incorrect decomposition; for example, agents may click to ``extract'' text instead of terminating with an already identified answer. Thus, TTED reduces observational complexity, while action generation remains sensitive to literal goal interpretations and environment dynamics beyond the base model's knowledge.

\section{Conclusion}

We introduced TTED, a label-free framework that learns simple test-time environments against compositional environment complexity. Across synthetic and realistic web benchmarks, TTED improves performance for both in-domain adaptation and cross-environment transfer. 
Our analyses further identify self-assessment quality and algorithm designs as key factors to effective test-time scaling. Ultimately, our findings establish that equipping LLM agents with environment decomposition skills is a critical step toward inference-time scalable real-world web automation.

\section{Ethical Statement}

The deployment of robust LLM web agents has significant potential to streamline digital workflows. By addressing the compositional generalization gap---a key barrier to transferring agents from controlled benchmarks to cluttered real-world interfaces---TTED could support more reliable autonomous digital assistants. Its label-free test-time adaptation also increases the potential for long-horizon behaviors that scale at inference time. Nevertheless, we advocate for the strictly monitored use of such autonomous agents. TTED remains a foundational prototype rather than a production-ready system, and broader safeguards are required against digital security risks and automated misuse.

\bibliography{aaai2027}

@String{Computing = "Computing" }

@String{Computer = "{IEEE} Computer" }

@String{Springer = "Springer-Verlag" }

@ArtifactSoftware{R,
    title = {R: A Language and Environment for Statistical Computing},
    author = {{R Core Team}},
    organization = {R Foundation for Statistical Computing},
    address = {Vienna, Austria},
    year = {2019},
    url = {https://www.R-project.org/},
}

@article{
furuta2024exposing,
title={Exposing Limitations of Language Model Agents in Sequential-Task Compositions on the Web},
author={Hiroki Furuta and Yutaka Matsuo and Aleksandra Faust and Izzeddin Gur},
journal={Transactions on Machine Learning Research},
issn={2835-8856},
year={2024},
url={https://openreview.net/forum?id=Y9kAsYIjYc},
note={}
}

@InProceedings{pmlr-v267-akyurek25a,
  title = 	 {The Surprising Effectiveness of Test-Time Training for Few-Shot Learning},
  author =       {Aky\"{u}rek, Ekin and Damani, Mehul and Zweiger, Adam and Qiu, Linlu and Guo, Han and Pari, Jyothish and Kim, Yoon and Andreas, Jacob},
  booktitle = 	 {Proceedings of the 42nd International Conference on Machine Learning},
  pages = 	 {942--963},
  year = 	 {2025},
  editor = 	 {Singh, Aarti and Fazel, Maryam and Hsu, Daniel and Lacoste-Julien, Simon and Berkenkamp, Felix and Maharaj, Tegan and Wagstaff, Kiri and Zhu, Jerry},
  volume = 	 {267},
  series = 	 {Proceedings of Machine Learning Research},
  month = 	 {13--19 Jul},
  publisher =    {PMLR},
  url = 	 {https://proceedings.mlr.press/v267/akyurek25a.html}
}

@article{yuksekgonul2026learningdiscovertesttime,
  title   = {Learning to Discover at Test Time},
  author  = {Yuksekgonul, Mert and Koceja, Daniel and Li, Xinhao 
             and Bianchi, Federico and McCaleb, Jed and Wang, Xiaolong 
             and Kautz, Jan and Choi, Yejin and Zou, James 
             and Guestrin, Carlos and Sun, Yu},
  journal = {arXiv preprint arXiv:2601.16175},
  year    = {2026}
}

@InProceedings{pmlr-v119-sun20b,
  title = 	 {Test-Time Training with Self-Supervision for Generalization under Distribution Shifts},
  author =       {Sun, Yu and Wang, Xiaolong and Liu, Zhuang and Miller, John and Efros, Alexei and Hardt, Moritz},
  booktitle = 	 {Proceedings of the 37th International Conference on Machine Learning},
  pages = 	 {9229--9248},
  year = 	 {2020},
  editor = 	 {III, Hal Daumé and Singh, Aarti},
  volume = 	 {119},
  series = 	 {Proceedings of Machine Learning Research},
  month = 	 {13--18 Jul},
  publisher =    {PMLR},
  url = 	 {https://proceedings.mlr.press/v119/sun20b.html}
}

@inproceedings{
zuo2025ttrl,
title={{TTRL}: Test-Time Reinforcement Learning},
author={Yuxin Zuo and Kaiyan Zhang and Li Sheng and Shang Qu and Ganqu Cui and Xuekai Zhu and Haozhan Li and Yuchen Zhang and Xinwei Long and Ermo Hua and Biqing Qi and Youbang Sun and Zhiyuan Ma and Lifan Yuan and Ning Ding and Bowen Zhou},
booktitle={The Thirty-ninth Annual Conference on Neural Information Processing Systems},
year={2025},
url={https://openreview.net/forum?id=VuVhgEiu20}
}

@article{liu2025agentenvironmentalignmentautomatedinterface,
  title={Agent-Environment Alignment via Automated Interface Generation},
  author={Liu, Kaiming and Lei, Xuanyu and Wang, Ziyue and Li, Peng and Liu, Yang},
  journal={arXiv preprint arXiv:2505.21055},
  year={2025}
}

@article{shao2024deepseekmathpushinglimitsmathematical,
  title={Deepseekmath: Pushing the limits of mathematical reasoning in open language models},
  author={Shao, Zhihong and Wang, Peiyi and Zhu, Qihao and Xu, Runxin and Song, Junxiao and Bi, Xiao and Zhang, Haowei and Zhang, Mingchuan and Li, YK and Wu, Yang and others},
  journal={arXiv preprint arXiv:2402.03300},
  year={2024}
}

@inproceedings{
zhou2024webarena,
title={WebArena: A Realistic Web Environment for Building Autonomous Agents},
author={Shuyan Zhou and Frank F. Xu and Hao Zhu and Xuhui Zhou and Robert Lo and Abishek Sridhar and Xianyi Cheng and Tianyue Ou and Yonatan Bisk and Daniel Fried and Uri Alon and Graham Neubig},
booktitle={The Twelfth International Conference on Learning Representations},
year={2024},
url={https://openreview.net/forum?id=oKn9c6ytLx}
}

@InProceedings{pmlr-v235-drouin24a,
  title = 	 {{W}ork{A}rena: How Capable are Web Agents at Solving Common Knowledge Work Tasks?},
  author =       {Drouin, Alexandre and Gasse, Maxime and Caccia, Massimo and Laradji, Issam H. and Del Verme, Manuel and Marty, Tom and Vazquez, David and Chapados, Nicolas and Lacoste, Alexandre},
  booktitle = 	 {Proceedings of the 41st International Conference on Machine Learning},
  pages = 	 {11642--11662},
  year = 	 {2024},
  editor = 	 {Salakhutdinov, Ruslan and Kolter, Zico and Heller, Katherine and Weller, Adrian and Oliver, Nuria and Scarlett, Jonathan and Berkenkamp, Felix},
  volume = 	 {235},
  series = 	 {Proceedings of Machine Learning Research},
  month = 	 {21--27 Jul},
  publisher =    {PMLR},
  url = 	 {https://proceedings.mlr.press/v235/drouin24a.html}
}

@inproceedings{
boisvert2024workarena,
title={WorkArena++: Towards Compositional Planning and Reasoning-based Common Knowledge Work Tasks},
author={L{\'e}o Boisvert and Megh Thakkar and Maxime Gasse and Massimo Caccia and Thibault Le Sellier de Chezelles and Quentin Cappart and Nicolas Chapados and Alexandre Lacoste and Alexandre Drouin},
booktitle={The Thirty-eight Conference on Neural Information Processing Systems Datasets and Benchmarks Track},
year={2024},
url={https://openreview.net/forum?id=PCjK8dqrWW}
}

@article{
chezelles2025the,
title={The BrowserGym Ecosystem for Web Agent Research},
author={Thibault Le Sellier de Chezelles and Maxime Gasse and Alexandre Lacoste and Massimo Caccia and Alexandre Drouin and L{\'e}o Boisvert and Megh Thakkar and Tom Marty and Rim Assouel and Sahar Omidi Shayegan and Lawrence Keunho Jang and Xing Han L{\`u} and Ori Yoran and Dehan Kong and Frank F. Xu and Siva Reddy and Graham Neubig and Quentin Cappart and Russ Salakhutdinov and Nicolas Chapados},
journal={Transactions on Machine Learning Research},
issn={2835-8856},
year={2025},
url={https://openreview.net/forum?id=5298fKGmv3},
note={Expert Certification}
}

@article{hu2025reinforcestabilizingcriticfreepolicy,
  title={Reinforce++: Stabilizing critic-free policy optimization with global advantage normalization},
  author={Hu, Jian and Liu, Jason Klein and Xu, Haotian and Shen, Wei},
  journal={arXiv preprint arXiv:2501.03262},
  year={2025}
}

@article{williams1992simple,
  title={Simple statistical gradient-following algorithms for connectionist reinforcement learning},
  author={Williams, Ronald J},
  journal={Machine learning},
  volume={8},
  pages={229--256},
  year={1992},
  publisher={Springer}
}

@inproceedings{10.1145/3689031.3696075,
author = {Sheng, Guangming and Zhang, Chi and Ye, Zilingfeng and Wu, Xibin and Zhang, Wang and Zhang, Ru and Peng, Yanghua and Lin, Haibin and Wu, Chuan},
title = {HybridFlow: A Flexible and Efficient RLHF Framework},
year = {2025},
isbn = {9798400711961},
publisher = {Association for Computing Machinery},
address = {New York, NY, USA},
url = {https://doi.org/10.1145/3689031.3696075},
doi = {10.1145/3689031.3696075},
booktitle = {Proceedings of the Twentieth European Conference on Computer Systems},
pages = {1279–1297},
numpages = {19},
location = {Rotterdam, Netherlands},
series = {EuroSys '25}
}

@inproceedings{
zheng2023judging,
title={Judging {LLM}-as-a-Judge with {MT}-Bench and Chatbot Arena},
author={Lianmin Zheng and Wei-Lin Chiang and Ying Sheng and Siyuan Zhuang and Zhanghao Wu and Yonghao Zhuang and Zi Lin and Zhuohan Li and Dacheng Li and Eric Xing and Hao Zhang and Joseph E. Gonzalez and Ion Stoica},
booktitle={Thirty-seventh Conference on Neural Information Processing Systems Datasets and Benchmarks Track},
year={2023},
url={https://openreview.net/forum?id=uccHPGDlao}
}

@article{singh2025openaigpt5card,
  title={Openai gpt-5 system card},
  author={Singh, Aaditya and Fry, Adam and Perelman, Adam and Tart, Adam and Ganesh, Adi and El-Kishky, Ahmed and McLaughlin, Aidan and Low, Aiden and Ostrow, AJ and Ananthram, Akhila and others},
  journal={arXiv preprint arXiv:2601.03267},
  year={2025}
}

@article{liu2025inferencetimescalinggeneralistreward,
  title={Inference-time scaling for generalist reward modeling},
  author={Liu, Zijun and Wang, Peiyi and Xu, Runxin and Ma, Shirong and Ruan, Chong and Li, Peng and Liu, Yang and Wu, Yu},
  journal={arXiv preprint arXiv:2504.02495},
  year={2025}
}

@article{zheng2025groupsequencepolicyoptimization,
  title={Group sequence policy optimization},
  author={Zheng, Chujie and Liu, Shixuan and Li, Mingze and Chen, Xiong-Hui and Yu, Bowen and Gao, Chang and Dang, Kai and Liu, Yuqiong and Men, Rui and Yang, An and others},
  journal={arXiv preprint arXiv:2507.18071},
  year={2025}
}

@inproceedings{
yang2025agentoccam,
title={AgentOccam: A Simple Yet Strong Baseline for {LLM}-Based Web Agents},
author={Ke Yang and Yao Liu and Sapana Chaudhary and Rasool Fakoor and Pratik Chaudhari and George Karypis and Huzefa Rangwala},
booktitle={The Thirteenth International Conference on Learning Representations},
year={2025},
url={https://openreview.net/forum?id=oWdzUpOlkX}
}

@article{lee2026agentictesttimescalingwebagents,
  title={Agentic test-time scaling for webagents},
  author={Lee, Nicholas and Erdogan, Lutfi Eren and John, Chris Joseph and Krishnapillai, Surya and Mahoney, Michael W and Keutzer, Kurt and Gholami, Amir},
  journal={arXiv preprint arXiv:2602.12276},
  year={2026}
}

@inproceedings{
shen2025thinking,
title={Thinking vs. Doing: Improving Agent Reasoning by  Scaling Test-Time Interaction},
author={Junhong Shen and Hao Bai and Lunjun Zhang and Yifei Zhou and Amrith Setlur and Shengbang Tong and Diego Caples and Nan Jiang and Tong Zhang and Ameet Talwalkar and Aviral Kumar},
booktitle={The Thirty-ninth Annual Conference on Neural Information Processing Systems},
year={2025},
url={https://openreview.net/forum?id=un1TRwNgiv}
}

@inproceedings{
deng2023mindweb,
title={Mind2Web: Towards a Generalist Agent for the Web},
author={Xiang Deng and Yu Gu and Boyuan Zheng and Shijie Chen and Samuel Stevens and Boshi Wang and Huan Sun and Yu Su},
booktitle={Thirty-seventh Conference on Neural Information Processing Systems Datasets and Benchmarks Track},
year={2023},
url={https://openreview.net/forum?id=kiYqbO3wqw}
}

@inproceedings{he-etal-2024-webvoyager,
    title = "{W}eb{V}oyager: Building an End-to-End Web Agent with Large Multimodal Models",
    author = "He, Hongliang  and
      Yao, Wenlin  and
      Ma, Kaixin  and
      Yu, Wenhao  and
      Dai, Yong  and
      Zhang, Hongming  and
      Lan, Zhenzhong  and
      Yu, Dong",
    editor = "Ku, Lun-Wei  and
      Martins, Andre  and
      Srikumar, Vivek",
    booktitle = "Proceedings of the 62nd Annual Meeting of the Association for Computational Linguistics (Volume 1: Long Papers)",
    month = aug,
    year = "2024",
    address = "Bangkok, Thailand",
    publisher = "Association for Computational Linguistics",
    url = "https://aclanthology.org/2024.acl-long.371/",
    doi = "10.18653/v1/2024.acl-long.371",
    pages = "6864--6890"
}

@inproceedings{wei-etal-2025-webagent,
    title = "{W}eb{A}gent-R1: Training Web Agents via End-to-End Multi-Turn Reinforcement Learning",
    author = "Wei, Zhepei  and
      Yao, Wenlin  and
      Liu, Yao  and
      Zhang, Weizhi  and
      Lu, Qin  and
      Qiu, Liang  and
      Yu, Changlong  and
      Xu, Puyang  and
      Zhang, Chao  and
      Yin, Bing  and
      Yun, Hyokun  and
      Li, Lihong",
    editor = "Christodoulopoulos, Christos  and
      Chakraborty, Tanmoy  and
      Rose, Carolyn  and
      Peng, Violet",
    booktitle = "Proceedings of the 2025 Conference on Empirical Methods in Natural Language Processing",
    month = nov,
    year = "2025",
    address = "Suzhou, China",
    publisher = "Association for Computational Linguistics",
    url = "https://aclanthology.org/2025.emnlp-main.401/",
    doi = "10.18653/v1/2025.emnlp-main.401",
    pages = "7909--7928",
    ISBN = "979-8-89176-332-6"
}

@article{wang2025uitars2technicalreportadvancing,
  title={Ui-tars-2 technical report: Advancing gui agent with multi-turn reinforcement learning},
  author={Wang, Haoming and Zou, Haoyang and Song, Huatong and Feng, Jiazhan and Fang, Junjie and Lu, Junting and Liu, Longxiang and Luo, Qinyu and Liang, Shihao and Huang, Shijue and others},
  journal={arXiv preprint arXiv:2509.02544},
  year={2025}
}

@article{openai2024openaio1card,
  title={{OpenAI} o1 System Card},
  author={Jaech, Aaron and Kalai, Adam and Lerer, Adam and Richardson, Adam and El-Kishky, Ahmed and Low, Aiden and Helyar, Alec and Madry, Aleksander and Beutel, Alex and Carney, Alex and others},
  journal={arXiv preprint arXiv:2412.16720},
  year={2024}
}

@article{openai2024gpt4ocard,
  title={GPT-4o System Card},
  author={Hurst, Aaron and Lerer, Adam and Goucher, Adam P and Perelman, Adam and Ramesh, Aditya and Clark, Aidan and Ostrow, AJ and Welihinda, Akila and Hayes, Alan and Radford, Alec and others},
  journal={arXiv preprint arXiv:2410.21276},
  year={2024}
}

@article{yang2025qwen3technicalreport,
  title={Qwen3 technical report},
  author={Yang, An and Li, Anfeng and Yang, Baosong and Zhang, Beichen and Hui, Binyuan and Zheng, Bo and Yu, Bowen and Gao, Chang and Huang, Chengen and Lv, Chenxu and others},
  journal={arXiv preprint arXiv:2505.09388},
  year={2025}
}

@article{deepseekai2025deepseekr1incentivizingreasoningcapability,
      title={DeepSeek-R1 incentivizes reasoning in LLMs through reinforcement learning}, 
      author={DeepSeek-AI},
  journal      = {Nature},
  year         = {2025},
  volume       = {645},
  number       = {7952},
  pages        = {633--638},
  month        = sep,
  doi          = {10.1038/s41586-025-09422-z}
}

@inproceedings{
lu2025agentrewardbench,
title={AgentRewardBench: Evaluating Automatic Evaluations of Web Agent Trajectories},
author={Xing Han L{\`u} and Amirhossein Kazemnejad and Nicholas Meade and Arkil Patel and Dongchan Shin and Alejandra Zambrano and Karolina Stanczak and Peter Shaw and Christopher Pal and Siva Reddy},
booktitle={Second Conference on Language Modeling},
year={2025},
url={https://openreview.net/forum?id=fQcUZMPIvu}
}

@inproceedings{10.5555/3495724.3495883,
author = {Brown, Tom B. and Mann, Benjamin and Ryder, Nick and Subbiah, Melanie and Kaplan, Jared and Dhariwal, Prafulla and Neelakantan, Arvind and Shyam, Pranav and Sastry, Girish and Askell, Amanda and Agarwal, Sandhini and Herbert-Voss, Ariel and Krueger, Gretchen and Henighan, Tom and Child, Rewon and Ramesh, Aditya and Ziegler, Daniel M. and Wu, Jeffrey and Winter, Clemens and Hesse, Christopher and Chen, Mark and Sigler, Eric and Litwin, Mateusz and Gray, Scott and Chess, Benjamin and Clark, Jack and Berner, Christopher and McCandlish, Sam and Radford, Alec and Sutskever, Ilya and Amodei, Dario},
title = {Language models are few-shot learners},
year = {2020},
isbn = {9781713829546},
publisher = {Curran Associates Inc.},
address = {Red Hook, NY, USA},
booktitle = {Proceedings of the 34th International Conference on Neural Information Processing Systems},
articleno = {159},
numpages = {25},
location = {Vancouver, BC, Canada},
series = {NIPS '20}
}

@inproceedings{dong-etal-2024-survey,
    title = "A Survey on In-context Learning",
    author = "Dong, Qingxiu  and
      Li, Lei  and
      Dai, Damai  and
      Zheng, Ce  and
      Ma, Jingyuan  and
      Li, Rui  and
      Xia, Heming  and
      Xu, Jingjing  and
      Wu, Zhiyong  and
      Chang, Baobao  and
      Sun, Xu  and
      Li, Lei  and
      Sui, Zhifang",
    editor = "Al-Onaizan, Yaser  and
      Bansal, Mohit  and
      Chen, Yun-Nung",
    booktitle = "Proceedings of the 2024 Conference on Empirical Methods in Natural Language Processing",
    month = nov,
    year = "2024",
    address = "Miami, Florida, USA",
    publisher = "Association for Computational Linguistics",
    url = "https://aclanthology.org/2024.emnlp-main.64/",
    doi = "10.18653/v1/2024.emnlp-main.64",
    pages = "1107--1128"
}

@inproceedings{
chae2025web,
title={Web Agents with World Models: Learning and Leveraging Environment Dynamics in Web Navigation},
author={Hyungjoo Chae and Namyoung Kim and Kai Tzu-iunn Ong and Minju Gwak and Gwanwoo Song and Jihoon Kim and Sunghwan Kim and Dongha Lee and Jinyoung Yeo},
booktitle={The Thirteenth International Conference on Learning Representations},
year={2025},
url={https://openreview.net/forum?id=moWiYJuSGF}
}

@inproceedings{
pan2024autonomous,
title={Autonomous Evaluation and Refinement of Digital Agents},
author={Jiayi Pan and Yichi Zhang and Nicholas Tomlin and Yifei Zhou and Sergey Levine and Alane Suhr},
booktitle={First Conference on Language Modeling},
year={2024},
url={https://openreview.net/forum?id=NPAQ6FKSmK}
}

@inproceedings{
he2025advancing,
title={Advancing Language Multi-Agent Learning with Credit Re-Assignment for Interactive Environment Generalization},
author={Zhitao He and Zijun Liu and Peng Li and Yi R. Fung and Ming Yan and Ji Zhang and Fei Huang and Yang Liu},
booktitle={Second Conference on Language Modeling},
year={2025},
url={https://openreview.net/forum?id=SoEmgM1ioC}
}

@article{Xi2025Rise,
  author  = {Xi, Zhiheng and Chen, Wenxiang and Guo, Xin and He, Wei and Ding, Yiwen and Hong, Boyang and Zhang, Ming and Wang, Junzhe and Jin, Senjie and Zhou, Enyu and Zheng, Rui and Fan, Xiaoran and Wang, Xiao and Xiong, Limao and Zhou, Yuhao and Wang, Weiran and Jiang, Changhao and Zou, Yicheng and Liu, Xiangyang and Yin, Zhangyue and Dou, Shihan and Weng, Rongxiang and Qin, Wenjuan and Zheng, Yongyan and Qiu, Xipeng and Huang, Xuanjing and Zhang, Qi and Gui, Tao},
  title   = {The rise and potential of large language model based agents: a survey},
  journal = {Science China Information Sciences},
  year    = {2025},
  volume  = {68},
  number  = {2},
  pages   = {121101},
  doi     = {10.1007/s11432-024-4222-0},
  url     = {https://doi.org/10.1007/s11432-024-4222-0}
}

@misc{OpenAI2026GPT54,
  author       = "{OpenAI}",
  year         = "2026",
  month        = mar,
  title        = "Introducing GPT-5.4",
  lastaccessed = "March 31, 2026",
  url          = "https://openai.com/index/introducing-gpt-5-4/",
}

@misc{liu-li-2025-rl-collapse,
  title = {When Speed Kills Stability: Demystifying {RL} Collapse from the Training-Inference Mismatch},
  author = {Liu, Jiacai and Li, Yingru and Fu, Yuqian and Wang, Jiawei and Liu, Qian and Shen, Yu},
  year = {2025},
  month = sep,
  lastaccessed = "March 31, 2026",
  url = {https://richardli.xyz/rl-collapse}
}

@inproceedings{
erdogan2025planandact,
title={Plan-and-Act: Improving Planning of Agents for Long-Horizon Tasks},
author={Lutfi Eren Erdogan and Hiroki Furuta and Sehoon Kim and Nicholas Lee and Suhong Moon and Gopala Anumanchipalli and Kurt Keutzer and Amir Gholami},
booktitle={Forty-second International Conference on Machine Learning},
year={2025},
url={https://openreview.net/forum?id=ybA4EcMmUZ}
}

@InProceedings{pmlr-v70-shi17a,
  title = 	 {World of Bits: An Open-Domain Platform for Web-Based Agents},
  author =       {Tianlin Shi and Andrej Karpathy and Linxi Fan and Jonathan Hernandez and Percy Liang},
  booktitle = 	 {Proceedings of the 34th International Conference on Machine Learning},
  pages = 	 {3135--3144},
  year = 	 {2017},
  editor = 	 {Precup, Doina and Teh, Yee Whye},
  volume = 	 {70},
  series = 	 {Proceedings of Machine Learning Research},
  month = 	 {06--11 Aug},
  publisher =    {PMLR},
  url = 	 {https://proceedings.mlr.press/v70/shi17a.html}
}

@inproceedings{
towers2025gymnasium,
title={Gymnasium: A Standard Interface for Reinforcement Learning Environments},
author={Mark Towers and Ariel Kwiatkowski and John U. Balis and Gianluca De Cola and Tristan Deleu and Manuel Goul{\~a}o and Kallinteris Andreas and Markus Krimmel and Arjun KG and Rodrigo De Lazcano Perez-Vicente and J K Terry and Andrea Pierr{\'e} and Sander V Schulhoff and Jun Jet Tai and Hannah Tan and Omar G. Younis},
booktitle={The Thirty-ninth Annual Conference on Neural Information Processing Systems Datasets and Benchmarks Track},
year={2025},
url={https://openreview.net/forum?id=qPMLvJxtPK}
}

@inproceedings{
yao2023react,
title={ReAct: Synergizing Reasoning and Acting in Language Models},
author={Shunyu Yao and Jeffrey Zhao and Dian Yu and Nan Du and Izhak Shafran and Karthik R Narasimhan and Yuan Cao},
booktitle={The Eleventh International Conference on Learning Representations },
year={2023},
url={https://openreview.net/forum?id=WE_vluYUL-X}
}

@inproceedings{
yao2022webshop,
title={WebShop: Towards Scalable Real-World Web Interaction with Grounded Language Agents},
author={Shunyu Yao and Howard Chen and John Yang and Karthik R Narasimhan},
booktitle={Advances in Neural Information Processing Systems},
editor={Alice H. Oh and Alekh Agarwal and Danielle Belgrave and Kyunghyun Cho},
year={2022},
url={https://openreview.net/forum?id=R9KnuFlvnU}
}

@inproceedings{
abedsoltan2025task,
title={Task Generalization with Autoregressive Compositional Structure: Can Learning from \$D\$ Tasks Generalize to \$D{\textasciicircum}T\$ Tasks?},
author={Amirhesam Abedsoltan and Huaqing Zhang and Kaiyue Wen and Hongzhou Lin and Jingzhao Zhang and Mikhail Belkin},
booktitle={Forty-second International Conference on Machine Learning},
year={2025},
url={https://openreview.net/forum?id=iZdGZSWe1A}
}

@inproceedings{song-etal-2025-beyond,
    title = "Beyond Browsing: {API}-Based Web Agents",
    author = "Song, Yueqi  and
      Xu, Frank F.  and
      Zhou, Shuyan  and
      Neubig, Graham",
    editor = "Che, Wanxiang  and
      Nabende, Joyce  and
      Shutova, Ekaterina  and
      Pilehvar, Mohammad Taher",
    booktitle = "Findings of the Association for Computational Linguistics: ACL 2025",
    month = jul,
    year = "2025",
    address = "Vienna, Austria",
    publisher = "Association for Computational Linguistics",
    url = "https://aclanthology.org/2025.findings-acl.577/",
    doi = "10.18653/v1/2025.findings-acl.577",
    pages = "11066--11085",
    ISBN = "979-8-89176-256-5"
}

@inproceedings{
wei2022chain,
title={Chain of Thought Prompting Elicits Reasoning in Large Language Models},
author={Jason Wei and Xuezhi Wang and Dale Schuurmans and Maarten Bosma and brian ichter and Fei Xia and Ed H. Chi and Quoc V Le and Denny Zhou},
booktitle={Advances in Neural Information Processing Systems},
editor={Alice H. Oh and Alekh Agarwal and Danielle Belgrave and Kyunghyun Cho},
year={2022},
url={https://openreview.net/forum?id=_VjQlMeSB_J}
}

@inproceedings{xu-etal-2025-crab,
    title = "{CRAB}: Cross-environment Agent Benchmark for Multimodal Language Model Agents",
    author = "Xu, Tianqi  and
      Chen, Linyao  and
      Wu, Dai-Jie  and
      Chen, Yanjun  and
      Zhang, Zecheng  and
      Yao, Xiang  and
      Xie, Zhiqiang  and
      Chen, Yongchao  and
      Liu, Shilong  and
      Qian, Bochen  and
      Yang, Anjie  and
      Jin, Zhaoxuan  and
      Deng, Jianbo  and
      Torr, Philip  and
      Ghanem, Bernard  and
      Li, Guohao",
    editor = "Che, Wanxiang  and
      Nabende, Joyce  and
      Shutova, Ekaterina  and
      Pilehvar, Mohammad Taher",
    booktitle = "Findings of the Association for Computational Linguistics: ACL 2025",
    month = jul,
    year = "2025",
    address = "Vienna, Austria",
    publisher = "Association for Computational Linguistics",
    url = "https://aclanthology.org/2025.findings-acl.1113/",
    doi = "10.18653/v1/2025.findings-acl.1113",
    pages = "21607--21647",
    ISBN = "979-8-89176-256-5"
}

@inproceedings{10.1145/3600006.3613165,
author = {Kwon, Woosuk and Li, Zhuohan and Zhuang, Siyuan and Sheng, Ying and Zheng, Lianmin and Yu, Cody Hao and Gonzalez, Joseph and Zhang, Hao and Stoica, Ion},
title = {Efficient Memory Management for Large Language Model Serving with PagedAttention},
year = {2023},
isbn = {9798400702297},
publisher = {Association for Computing Machinery},
address = {New York, NY, USA},
url = {https://doi.org/10.1145/3600006.3613165},
doi = {10.1145/3600006.3613165},
booktitle = {Proceedings of the 29th Symposium on Operating Systems Principles},
pages = {611–626},
numpages = {16},
location = {Koblenz, Germany},
series = {SOSP '23}
}

@inproceedings{reimers-gurevych-2019-sentence,
    title = "Sentence-{BERT}: Sentence Embeddings using {S}iamese {BERT}-Networks",
    author = "Reimers, Nils  and
      Gurevych, Iryna",
    editor = "Inui, Kentaro  and
      Jiang, Jing  and
      Ng, Vincent  and
      Wan, Xiaojun",
    booktitle = "Proceedings of the 2019 Conference on Empirical Methods in Natural Language Processing and the 9th International Joint Conference on Natural Language Processing (EMNLP-IJCNLP)",
    month = nov,
    year = "2019",
    address = "Hong Kong, China",
    publisher = "Association for Computational Linguistics",
    url = "https://aclanthology.org/D19-1410/",
    doi = "10.18653/v1/D19-1410",
    pages = "3982--3992"
}

@inproceedings{10.5555/3495724.3496209,
author = {Wang, Wenhui and Wei, Furu and Dong, Li and Bao, Hangbo and Yang, Nan and Zhou, Ming},
title = {MINILM: deep self-attention distillation for task-agnostic compression of pre-trained transformers},
year = {2020},
isbn = {9781713829546},
publisher = {Curran Associates Inc.},
address = {Red Hook, NY, USA},
booktitle = {Proceedings of the 34th International Conference on Neural Information Processing Systems},
articleno = {485},
numpages = {13},
location = {Vancouver, BC, Canada},
series = {NIPS '20}
}

@inproceedings{hong2024metagpt,
      title="Meta{GPT}: Meta Programming for A Multi-Agent Collaborative Framework",
      author="Sirui Hong and Mingchen Zhuge and Jonathan Chen and Xiawu Zheng and Yuheng Cheng and Jinlin Wang and Ceyao Zhang and Zili Wang and Steven Ka Shing Yau and Zijuan Lin and Liyang Zhou and Chenyu Ran and Lingfeng Xiao and Chenglin Wu and J{\"u}rgen Schmidhuber",
      booktitle="The Twelfth International Conference on Learning Representations",
      year="2024",
      url="https://openreview.net/forum?id=VtmBAGCN7o",
      
}

@inproceedings{liu2018reinforcement,
 author = {Evan Zheran Liu and Kelvin Guu and Panupong Pasupat and Tianlin Shi and Percy Liang},
 title = {Reinforcement Learning on Web Interfaces using Workflow-Guided Exploration},
 booktitle = {International Conference on Learning Representations ({ICLR})},
 url = {https://arxiv.org/abs/1802.08802},
 year = {2018},
}

@misc{selenium,
 title = "Selenium Webdriver",
 author = "Selenium",
 url = "https://www.selenium.dev/documentation/webdriver/",
 year = "2026",
}

@misc{OpenAI2026GPT55,
  author       = "{OpenAI}",
  year         = "2026",
  month        = apr,
  title        = "Introducing GPT-5.5",
  lastaccessed = "July 2, 2026",
  url          = "https://openai.com/index/introducing-gpt-5-5/",
}

@article{kerboua2025focusagent,
  title={Focusagent: Simple yet effective ways of trimming the large context of web agents},
  author={Kerboua, Imene and Shayegan, Sahar Omidi and Thakkar, Megh and L{\`u}, Xing Han and Boisvert, L{\'e}o and Caccia, Massimo and Espinas, J{\'e}r{\'e}my and Aussem, Alexandre and Eglin, V{\'e}ronique and Lacoste, Alexandre},
  journal={arXiv preprint arXiv:2510.03204},
  year={2025}
}

\appendix
\setcounter{secnumdepth}{2}

\newpage

\section{Discussions, Limitations, and Additional Related Work}
\label{app:dis & limit}
\subsection{Discussions}
\label{app:discuss}
\paragraph{\textbf{Compositional Generalization based on Environment Decomposition vs. Planning}}
We further discuss two distinct routes to achieving compositional: one through planning, which decomposes a complex goal into a sequence of sub-goals, and the other through environment decomposition, which structures the environment into simpler sub-environments. Prior work on long-horizon decision-making has predominantly followed the former paradigm, and has shown strong effectiveness across a wide range of domains~\citep{hong2024metagpt, yao2023react, erdogan2025planandact}. However, planning-based methods differ fundamentally from our approach of compositional generalization based on environment decomposition in several key aspects:

\subparagraph{(1) Difference in the level of decomposition:} Planning primarily operates at the level of goal decomposition, whereas TTED decomposes the entire environment, including goals, observations, and unique rewards. Formally, planning decomposes a global goal $G$ into a sequence of sub-goals,
$$G=(g_1,g_2,...,g_n)$$
where each $g_i$ represents a sub-goal to be achieved. In contrast, environment decomposition partitions the original environment into a collection of sub-environments,
\[
\mathcal{E} = (\mathcal{E}^{(1)}, \mathcal{E}^{(2)}, \ldots, \mathcal{E}^{(n)})
\]
where each sub-environment $\mathcal{E}_i$ encapsulates its own goal, observation space, and reward function, which can be formalized as $\mathcal{E}^{(i)}
=
(\mathcal{S}^{(i)},\mathcal{A},
 \mathcal{T},\mathcal{R}^{(i)})$. 

\subparagraph{(2) Difference in the optimization objective:}
Planning-based approaches typically decompose a complex task into a sequence of sub-goals. However, such approaches still operate in the original complex environment, where the optimization objective remains the reward of the full environment:
$$
\pi_{\mathrm{ori}}^*
=
\arg\max_{\pi}
\mathbb{E}_{\tau\sim\pi}
[\mathcal{R}(\tau)].
$$
In complex environments, estimating reward signals at test time is often expensive and inaccurate. 
To mitigate this issue, TTED explicitly optimizes the accumulated rewards across multiple sub-environments 

$$
\pi_{\mathrm{TTED}}^*
=
\arg\max_{\pi}
\sum_{i=1}^{n}
\mathbb{E}_{\tau^{(i)}\sim\pi}
[\mathcal{R}^{(i)}(\tau^{(i)})].
$$

Through environment decomposition, we obtain simplified sub-environments with reduced complexity and clearer objectives. In such simplified environments, more accurate self-assessment reward signals can be obtained, as the agent only needs to reason over simplified sub-goals and sub-observations. This advantage is further validated through human evaluation in Ablation Study section.

In summary, these differences highlight that environment decomposition is not merely an alternative implementation of planning, but a distinct paradigm that achieves compositional generalization.

\paragraph{\textbf{Synthetic Environment vs. Real-world Environment}}

Synthetic environments are typically constructed by composing multiple simpler environments into a new, more complex one~\citep{furuta2024exposing, abedsoltan2025task}. Through such composition, these environments naturally exhibit multi-step structure and increased goal and observation complexity. A representative example is CompWoB~\citep{furuta2024exposing}, where each environment is formed by combining several MiniWoB~\citep{liu2018reinforcement} environments. Formally, a CompWoB environment can be expressed as
\[
\mathcal{E} = (\mathcal{E}_1, \mathcal{E}_2, \ldots, \mathcal{E}_n)
\]
where $\mathcal{E}$ denotes the composite CompWoB environment and each $\mathcal{E}_i$ corresponds to an individual MiniWoB environment. Due to their compositional construction, synthetic environments exhibit clear internal boundaries between sub-environments. As a result, environment decomposition is naturally well-defined. This property makes synthetic environments particularly suitable for studying compositional generalization under controlled settings. 

In contrast, real-world environments lack such explicit compositional structure. Internal boundaries between each decomposition part are often implicit, overlapping, or ambiguous, making it difficult to obtain clear and well-defined environment decompositions. In these settings, environment decomposition cannot be trivially derived from the environment construction, but instead relies heavily on the agent capabilities of planning and reasoning to infer meaningful sub-environments. Also, for practical deployment, unsafe operations or actions with critical side effects in sub-environemnt exploration should be suspended and wait for user confirmation. 

Consequently, although synthetic environments provide a convenient and controlled testbed for studying compositional generalization, the substantial gap between synthetic and real-world settings necessitates validating such methods in realistic environments. In real-world scenarios, compositional structure must be actively discovered rather than given by design, making it essential to evaluate whether improvements observed in synthetic benchmarks can translate to practical and complex environments. 

\paragraph{\textbf{TTED vs. Observation Pruning or Subgoal-driven RL}}
Observation pruning methods, such as AgentOccam~\citep{yang2025agentoccam} and FocusAgent~\citep{kerboua2025focusagent}, aim to manage long-context inputs by truncating, filtering, or selecting relevant observations from temporal trajectories. These methods are orthogonal to TTED: they primarily reduce the context length at inference time, whereas TTED uses localized environment decomposition as \emph{learning signals} for label-free test-time adaptation. In this sense, prior observation-reduction methods motivate the need for more effective self-assessment and adaptation mechanisms, rather than weakening our setting.

Subgoal-driven RL also differs from TTED in both motivation and usage. Methods that use checklists, principles, or subgoals for reward modeling typically aim to enhance the overall reward signal by aggregating multiple intermediate scores along a trajectory. In contrast, TTED decomposes a complex task into a \emph{specific sub-goal} aligned with the corresponding decomposed environment, enabling the agent to perform localized self-learning during test-time adaptation. Therefore, TTED is not designed to provide a full-trajectory outcome reward, but to construct localized learning signals that better match the current sub-environment.

\subsection{Limitations}
\label{app:limit}
Despite the effectiveness of our approach, it has several limitations that merit discussion.

\paragraph{The quality of environment decomposition is inherently dependent on the capability of pre-trained models.}
As discussed in Appendix~\ref{app:discuss}, TTED relies on the ability of the backbone model to identify meaningful sub-environments from complex environments. When the decomposition quality is poor due to extreme incompetence of models, errors or ambiguities in the resulting sub-environments may propagate to downstream decision-making, limiting the effectiveness of compositional generalization. This issue can be pronounced in realistic settings, where environment boundaries are more implicit and more difficult to infer. Improving the robustness and reliability of environment decomposition remains an important direction for future work.

\paragraph{Agent self-assessment in multi-turn web environments is designed with simplicity and does not leverage sources outside the backbone model.}
TTED employs agent self-assessment to provide reward signals that assess action execution and decomposition quality in the test time. While this design enables flexible and scalable training without ground-truth supervision, it also introduces potential bias and noise inherited from the backbone model itself. Inaccurate or inconsistent judgments may affect training stability and limit performance gains. Utilizing larger backbone models or exploring more advanced self-rewarding approaches beyond GRM, e.g., script-based simulator for static webpages in TTED with ICL, could help mitigate this limitation in future. 

\subsection{Additional Background on LLM Web Agents and Evaluation}
\label{app:related_work}
Recent advancements have positioned LLMs as capable agents for sequential decision-making~\citep{yao2023react}, and thus general web automation is implemented as a natural application domain for agents to interact with complex observation spaces, including HTML, DOM trees, accessibility metadata, screenshots, or mixed multimodal representations, and multi-level action interfaces, including low-level clicks and keystrokes, high-level API calls, or script-based code generation~\citep{he-etal-2024-webvoyager,song-etal-2025-beyond,wei-etal-2025-webagent,yang2025agentoccam}, and perform diverse tasks such as basic component manipulation, web shopping, information retrieval, and more realistic scenarios~\citep{pmlr-v70-shi17a,yao2022webshop,deng2023mindweb,zhou2024webarena,pmlr-v235-drouin24a}. Recent efforts have explored training LLMs with long reasoning capabilities~\citep{openai2024openaio1card,deepseekai2025deepseekr1incentivizingreasoningcapability} on interactive web environments with RL~\citep{wei-etal-2025-webagent,wang2025uitars2technicalreportadvancing,shen2025thinking,OpenAI2026GPT54}. For evaluation, \citet{chezelles2025the} developed a unified protocol for benchmarking web agents, while \citet{lu2025agentrewardbench} consider how well LLMs can judge web agent performance as reward models.

\section{Additional Experimental Results}\label{app:add_exp}

\subsection{CompWob+ Results with Larger LLMs}
\label{app:additional_compwob}
\begin{figure}[h]
  \centering
  \includegraphics[width=0.6\linewidth]{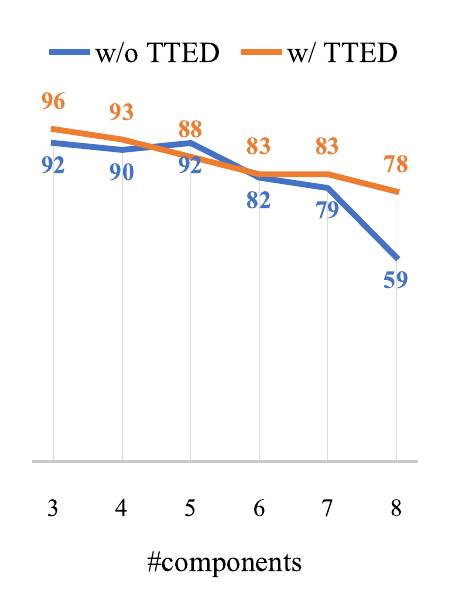}
  \caption{Task success rates (\%) on CompWob+ with Qwen3 14B}
  \label{fig:compwob_qwen_14b}
\end{figure}
To further evaluate the robustness of TTED across different models, we additionally conduct experiments on CompWoB+ using Qwen3-14B. The results are shown in Figure~\ref{fig:compwob_qwen_14b}. 

As model size increases, the base model itself demonstrates improved performance across all compositional tasks. However, the model still experiences a significant performance drop when the composition number reaches 8, with a 25.3\% decrease compared to tasks with 7 compositions. In contrast, with TTED combined with ICL, the performance remains stable as task complexity increases, achieving 78\% accuracy even for tasks with 8 compositions. These results demonstrate that TTED remains effective when applied to different models.

\subsection{Webarena Results with Additional LLMs}
\label{app:additional_webarena}
To test 10B-scale open-sourced models with long CoT reasoning ability, we train Qwen3-14B on WebArena with RL and evaluate it using the same protocol as Qwen3-8B. The results are shown in Table~\ref{table:expand_webarena}.

\begin{table}[h]
\centering
\resizebox{0.65\linewidth}{!}{
\begin{tabular}{llc}
\toprule
\textbf{Model} & \textbf{Method} & \textbf{WebArena SR} \\
\midrule
\textcolor{gray}{Gemma3-12B} & \textcolor{gray}{TTI} & \textcolor{gray}{22.1} \\
\textcolor{gray}{GPT-4o} & \textcolor{gray}{-} & \textcolor{gray}{23.5} \\
\textcolor{gray}{GPT-5.5*} & \textcolor{gray}{-} & \textcolor{gray}{67.3} \\
\midrule
Qwen3-8B & - & 12.1 \\
& RL w/ GT & 14.8 \\
& TTT (RL) & 12.5 \\
& \textbf{TTED (RL)} & \textbf{21.0} \\
\midrule
Qwen3-14B & - & 19.3 \\
& RL w/ GT & 20.0 \\
& TTT (RL) & 18.7 \\
& \textbf{TTED (RL)} & \textbf{22.5} \\
\bottomrule
\end{tabular}
}
\caption{Success rate (SR, \%) on the selected WebArena subset. Results in gray are provided as contextual references only. TTI is not a test-time training method and uses ground-truth labels in the benchmark. GPT-5.5* is evaluated on full WebArena-Verified~\citep{OpenAI2026GPT55}, which shares similar task difficulty but uses different verifiers.}
\label{table:expand_webarena}
\vspace{-0.5em}
\end{table}

Compared to bare Qwen3-14B, TTED achieves relative gains of 16.6\% on undecomposed environments and 4.2\% on decomposed environments, while still outperforming ground-truth RL. The smaller margin may reflect higher absolute performance and increasingly dominant bottlenecks beyond environment complexity; other baselines follow the Qwen3-8B trend.

In addition, we evaluate the performance of proprietary models on WebArena, where GPT-4o and GPT-5.5 achieve success rates of 23.5\% and 67.3\%, respectively. Although TTED does not match the performance of these proprietary models, model scale should also be taken into account. For the same Qwen3-8B backbone, TTED improves WebArena from 12.1\% to 21.0\% and WorkArena from 18.2\% to 22.7\% (73.6\% and 24.7\% relative improvement, respectively).

\subsection{Statistic Significance Tests}
\label{app:significance}
We repeated the main WebArena results (Table~1, Figure~4) three times. Rollout sampling uses temperature 1.0, whereas evaluation uses temperature 0. Across runs, the absolute variation is below 0.5\% and thus results are not changed. Over the 561 WebArena tasks, Qwen3-8B TTED achieves an SR of 21.0\% with a 95\% Wilson CI of $[17.9, 24.6]$, compared with $[9.7, 15.1]$ for the base model, $[12.1, 18.0]$ for RL w/ GT, and $[10.0, 15.5]$ for TTT. One-sided aggregate proportion tests show that TTED significantly outperforms these baselines ($p<10^{-4}$, $p=0.0032$, and $p<10^{-4}$, respectively). For Qwen3-14B, the gains remain positive ($p=0.093$, $p=0.153$, and $p=0.061$, respectively) (Table~\ref{table:expand_webarena}). 

\begin{table}[h]
  \caption{Efficiency Measures of TTED-trained models on WebArena. ``\#Steps'' denotes the average number of interaction steps in a task, and ``\#Prompt Tokens'' denotes the average input tokens of the action generation policy.}
  \label{tab:token}
  \resizebox{\linewidth}{!}{
  \begin{tabular}{lrccc}
    \toprule
    \textbf{Method} & \textbf{\#Sub-Envs} & \textbf{WebArena SR} (\%) & \textbf{\#Steps} & \textbf{\#Prompt Tokens} \\
    \midrule
    Qwen3-8B & 0  & 12.1 & 4.72 & 6338.9 \\
    TTED (RL) & 3200& 21.0 & 8.29 & 3948.2 \\
  \bottomrule
\end{tabular}
  }

\end{table}

\subsection{Efficiency Analysis}
\label{app:efficiency}
Table~\ref{tab:token} demonstrates TTED's context efficiency. While exploratory trial steps naturally increase the average interaction steps per task (8.29 vs. 4.72), restricting the action generation policy to partitioned sub-observations significantly reduces average prompt token consumption. In summary, TTED extends interaction scales during test-time learning, and localized observations successfully reduce per-action computational overhead, which is very critical for real-world web automation where textual representation of websites can be six-figure token counts. 

\section{Qualitative Studies}\label{app:qual}

In this section, we provide detailed qualitative case studies to complement the observations discussed in Experiments section. Including cases from CompWob+~\citep{furuta2024exposing} and WebArena~\citep{zhou2024webarena}.

\subsection{Webarena Case Studies}
\label{app:case_webarena}
To further understand the behavior of TTED, we present three representative case studies from WebArena. These cases illustrate (1) success enabled by TTED, (2) performance improvement after test-time training, and (3) failure cases where correct decomposition still leads to errors due to action-space misalignment.

\paragraph{Case 1: Success with TTED but failure without TTED}  
In this case, the agent without TTED fails due to misinterpretation of key elements in the complex environment. In contrast, the agent with TTED isolates relevant components, correctly identifies the key elements, and successfully completes the task. This case demonstrates that environment decomposition simplifies the environment, enabling agents to perform more accurate reasoning and reducing hallucinations.

\begin{figure}[h]
  \centering
  \includegraphics[width=\linewidth]{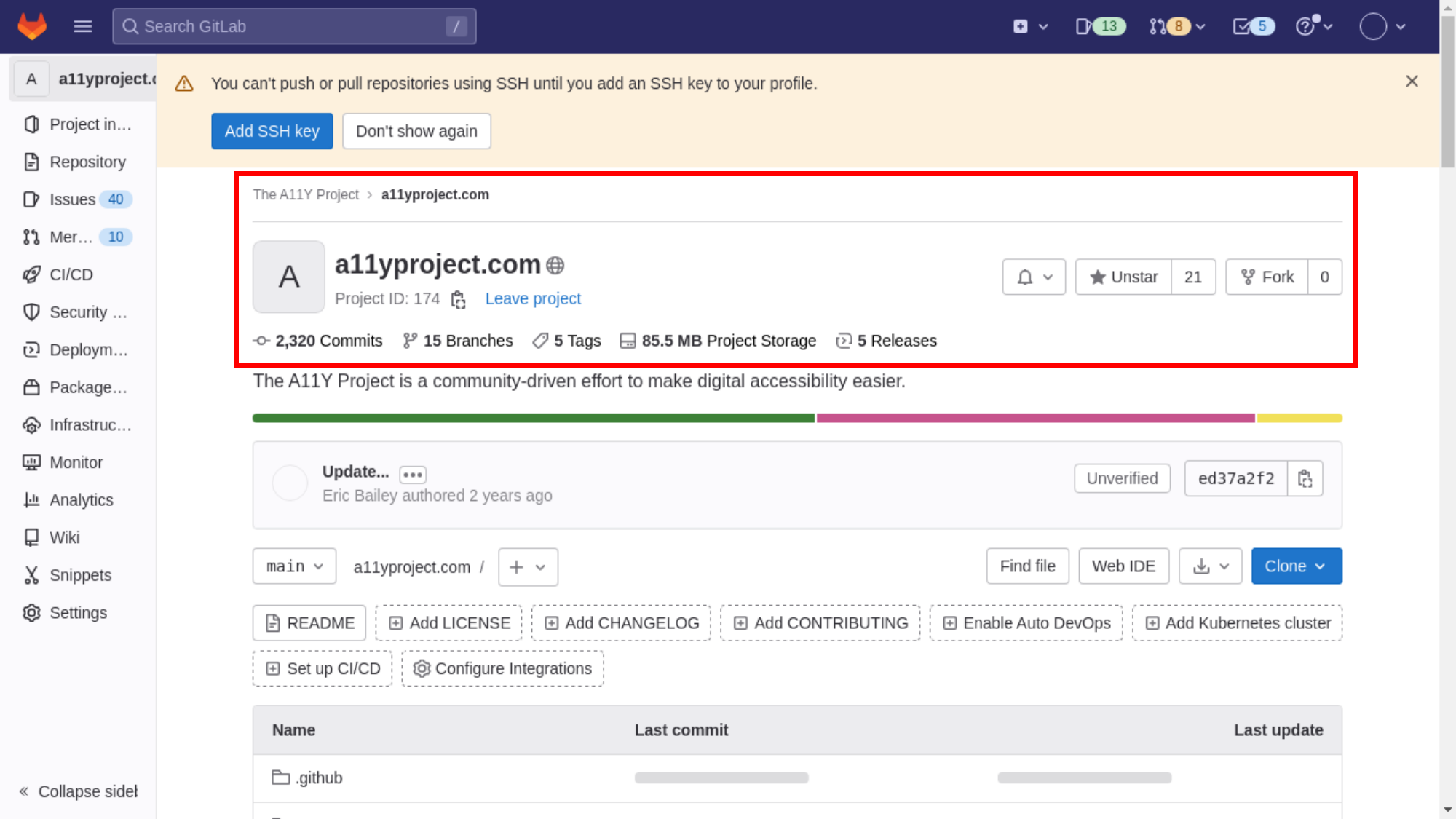}
  \caption{Observation for \# Case 1.}
  \label{fig:qual_case_1_obs}
\end{figure}

\begin{figure*}[t]
  \includegraphics[width=\textwidth]{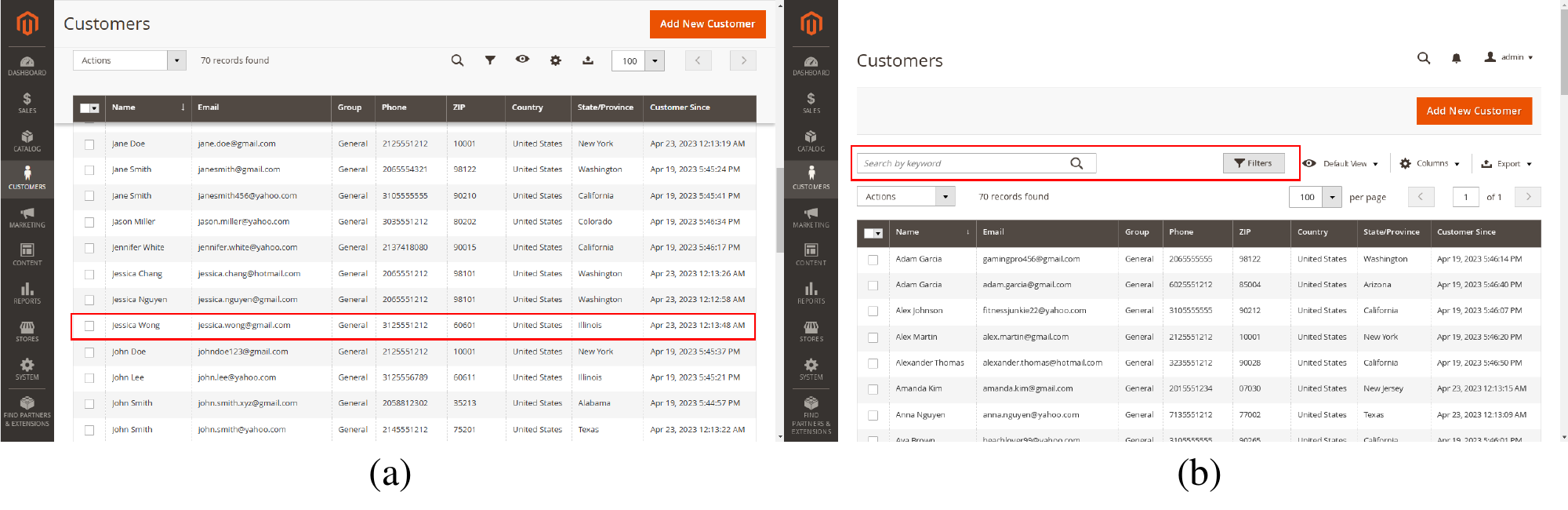}
  \caption{Observation for \# Case 2.}
  \label{fig:qual_case_2_obs}
\end{figure*}

\begin{casebox}{\# Case 1: Success with TTED vs. Failure without TTED}

\textbf{Goal:} How many commits did Eric make to a11yproject on 3/2?

\textbf{Website:} GitLab 

\textbf{Common Steps:}

- step 1. click(``393'')    \# navigate to ``The A11Y Project / a11yproject.com'' repository page

\textbf{Current Page:} Repository page, see Figure~\ref{fig:qual_case_1_obs} for detail.

\textbf{Observation:} 

RootWebArea `The A11Y Project / a11yproject.com · GitLab', focused

	\hspace{2em}[67] banner `'
    
		\hspace{4em}[68] link `Skip to content'
        
		\hspace{4em}StaticText `GitLab'
        
		\hspace{4em}[74] link `Dashboard'
        
			\hspace{6em}[75] image `'

            \hspace{6em}...

\textbf{Divergence Point:}

---------------------------------------------------------------------------------------------------

\textcolor{red}{\# failure in w/o decomposition model}

\textcolor{blue}{<think>}

\textcolor{blue}{1. The agent think `commit history’ should be found in current page.}

First, I need to look at the current web page's accessibility tree provided… The observation includes various elements, but \textcolor{red!70}{the main area of interest is the commit history.}

…

\textcolor{blue}{2. Then, the agent interpreted the current page as merely a file list, without identifying any elements related to commit history.}

Looking at the elements, there's a table labeled `Files, directories, and submodules in the path / for commit reference main’… But in the current observation, \textcolor{red!70}{there's no explicit mention of a commit history table or a list of commits.} The table here seems to list files and directories, not commits.

...

\textcolor{blue}{3. The agent identified the key element required to complete the task, but mistakenly interpreted it as plain text rather than a link, and therefore excluded it.}

Wait, there's a \textcolor{red!70}{StaticText ``2,320 Commits''} in the listitem under the navigation. \textcolor{red!70}{That might indicate that there are 2,320 commits in total, but that's not specific to Eric or March 2nd. }

\textcolor{blue}{</think>}

\textcolor{blue}{Final Action:} report\_infeasible(``The current page does not provide commit history filtered by author or date. Eric's commits on March 2nd cannot be determined from the available data.'')
\hspace{1em}
\raisebox{-0.4em}{\includegraphics[height=1.2em]{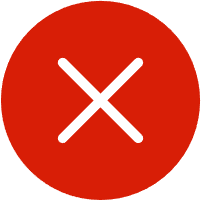}}

---------------------------------------------------------------------------------------------------

\textcolor{red}{\# success in w/ decomposition model}

\textcolor{blue}{- Goal Decomposition:} Click the link labeled `2,320 Commits' to view the project's commit history and identify commits made by Eric on March 2nd.

\textcolor{blue}{\ \ - Corresponding Reasoning:}
\textcolor{blue}{<think>}

\textcolor{blue}{1. Agent correctly interpreted the current page.}

\textcolor{red!70}{First, the agent is on the project's main page.}

...

\textcolor{blue}{2. Then, the agent identified the key sub-goal for the next step and proposed a potential solution.}

The next step would be to check if there's a way to view commits by a specific author and date.

...

- On GitLab, typically, you can \textcolor{red!70}{navigate to the repository's ``Repository'' section, then go to ``Commits'' or ``Activity'' to see commit history.}

- Alternatively, there might be a ``Commits'' link in the main content. The AXTree has a list under the main area with \textcolor{red!70}{``2,320 Commits'' as a link.}

\textcolor{blue}{3. Finally, the agent selected the second approach and ultimately formulated a concrete sub-goal.}
The AXTree has \textcolor{red!70}{a link labeled ``2,320 Commits''} under the main area's navigation. \textcolor{red!70}{Clicking that link would take the agent to the commit history page.}

\textcolor{blue}{</think>}

\textcolor{blue}{- Observation Decomposition:} See the region highlighted by the red rectangle in Figure~\ref{fig:qual_case_1_obs} for detail. 

\textcolor{blue}{Final Action:} click(`683')   \# Correctly click the ``2,320 Commits'' link.
\hspace{1em}
\raisebox{-0.4em}{\includegraphics[height=1.2em]{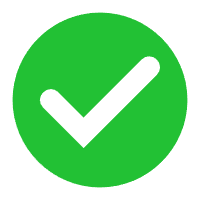}}

\end{casebox}

\paragraph{Case 2: Success after training.}  
For the same task, the agent before training adopts a brute-force strategy to scan all entries, which leads to hallucinations under complex contexts and ultimately results in failure. After training, the agent learns to take advantage of the search functionality to efficiently locate the target, leading to successful completion of the task. The key difference lies in the selection of the agent strategy and robustness under complex observations.

\begin{casebox}{\# Case 2: Success after Training vs. Failure before Training}

\textbf{Goal:} Find the customer name and email with phone number 2137418080

\textbf{Website:} CMS

\textbf{Common Steps: }

- step 1 \& 2. Navigate to `All customers' page

\textbf{Current Page:} Customers information page, see Figure~\ref{fig:qual_case_2_obs}(a) for detail.

\textbf{Observation:} 

Accessibility Tree:

RootWebArea `Customers / Customers / Magento Admin', focused

	\hspace{2em}[175] link `Magento Admin Panel'
    
		\hspace{4em}[176] image `Magento Admin Panel'
        
	\hspace{2em}[177] navigation `'
    
		\hspace{4em}[178] menubar `', orientation=`horizontal'
        
			\hspace{6em}[180] link `DASHBOARD'
            
				\hspace{8em}StaticText `'
                
				\hspace{8em}StaticText `DASHBOARD'
                
			\hspace{6em}[183] link `SALES'
            
				\hspace{8em}StaticText `'
                
				\hspace{8em}StaticText `SALES'

                \hspace{8em}...

\textbf{Divergence Point:} 

---------------------------------------------------------------------------------------------------

\textcolor{red}{\# failure in w/o train model}

\textcolor{blue}{- Goal Decomposition:} Click the row containing the phone number `2137418080' to select the corresponding customer record.

\textcolor{blue}{\ \ - Corresponding Reasoning:}
\textcolor{blue}{<think>}

\textcolor{blue}{1. The agent correctly interpreted the current page.}

Okay, let's see. The task is to find the customer name and email with the phone number 2137418080. \textcolor{red!70}{The current page is the Magento Admin's Customers section, and there's a table listing various customers with their details.}

...

\textcolor{blue}{2. Then, the agent chose to brute-force all entries instead of using the search function, which set the stage for subsequent errors.}

Looking at the AXTree, there's a table with columns like Name, Email, Phone, etc. The phone numbers are in the `Phone' column. \textcolor{red!70}{I need to scan through the rows to find the entry where the Phone column matches 2137418080.}

\textcolor{blue}{3. Due to the overly complex environment, the agent exhibited hallucination while searching for the target, ultimately leading to an incorrect answer.}

Wait, looking further down, there's a row \textcolor{red!70}{[2310] where the Phone is `2137418080'}. That's the one we need. 

\textcolor{red!70}{* Element [2310] corresponds to the entry with the phone number `2065551212', rather than the goal target.}

\textcolor{blue}{</think>}

\textcolor{blue}{- Observation Decomposition:} See the region highlighted by the red rectangle in Figure~\ref{fig:qual_case_2_obs}(a) for detail.  \textcolor{red!70}{\# The agent again hallucinated, modifying the entry with phone number `3035551212' into the supposed target entry.}

\textcolor{red!70}{* Hallucination: [2279] gridcell `3035551212'  \textbf{→}  \sout{[2279] gridcell `2137418080'} }
\hspace{1em}
\raisebox{-0.4em}{\includegraphics[height=1.2em]{figs/cross.png}}

\textcolor{blue}{\ \ - Corresponding Reasoning:}
\textcolor{blue}{<think>}

\textcolor{red!70}{… Yes, that's the row with the phone number `2137418080'. So the entire row [2268] and its children are relevant. …}

\textcolor{blue}{</think>}

\textcolor{blue}{Final Action:} click(`2279')  \textcolor{red!70}{\# The element [2279] actually is `3035551212'.}
\hspace{1em}
\raisebox{-0.4em}{\includegraphics[height=1.2em]{figs/cross.png}}

---------------------------------------------------------------------------------------------------

\textcolor{red}{\# success in w/ train model}

\textcolor{blue}{- Goal Decomposition:} Search for the phone number `2137418080' in the `Search by keyword' textbox and click the `Search' button to filter the customer list.

\textcolor{blue}{\ \ - Corresponding Reasoning:}
\textcolor{blue}{<think>}

\textcolor{blue}{1. The agent correctly interpreted the current page and proposed using search to locate the target.}

The current page is the Magento Admin's Customers section, and there's a table listing all customers. The user has already navigated to the Customers section, so now \textcolor{red!70}{they need to search through the customer data.}

\textcolor{blue}{2. Then, the agent confirms that the search box exists on the current page.}

\textcolor{red!70}{The next step should be to search for the phone number 2137418080 in the table.} ... .Wait, in the AXTree, \textcolor{red!70}{there's a textbox labeled `Search by keyword' and a button `Search'.} So maybe the agent should use that search functionality to filter the customers by phone number.

\textcolor{blue}{</think>}

\textcolor{blue}{- Observation Decomposition:} See the region highlighted by the red rectangle in Figure~\ref{fig:qual_case_2_obs}(b) for detail.  \textcolor{red!70}{\# The agent successfully locates the search box.}

\textcolor{blue}{Final Action:} fill(``854'', ``2137418080'')  \textcolor{red!70}{\# [854] textbox `Search by keyword'}
\hspace{1em}
\raisebox{-0.4em}{\includegraphics[height=1.2em]{figs/right.png}}

\end{casebox}

\paragraph{Case 3: Failure despite correct environment decomposition.}
In this case, the environment decomposition is correct, but the agent still fails due to action-space misalignment. Specifically, the agent overthinks when attempting to map the sub-goal to an unavailable “extract” action, ultimately selecting an incorrect action despite already identifying the correct answer.

\begin{figure}[h]
  \includegraphics[width=\linewidth]{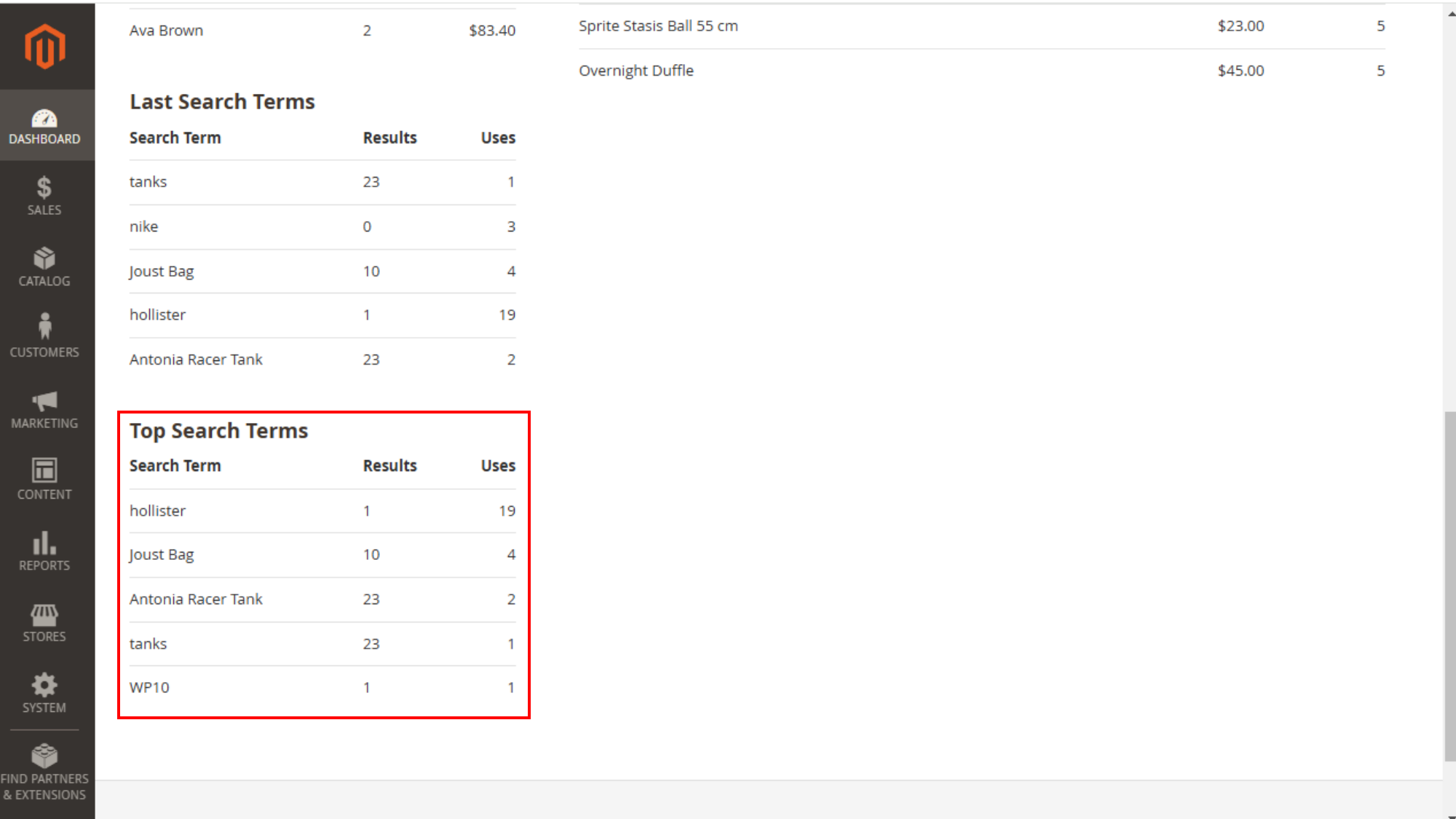}
  \caption{Observation for \# Case 3.}
  \label{fig:qual_case_3_obs}
\end{figure}

\begin{casebox}{\# Case 3: Failure Caused by Overthinking}

\textbf{Goal:} List the top 1 search terms in my store

\textbf{Website:} CMS

\textbf{Current Page:} Dashboard, see Figure~\ref{fig:qual_case_3_obs} for detail.

\textbf{Observation:}

RootWebArea `Dashboard / Magento Admin', focused

	\hspace{2em}[148] link `Magento Admin Panel'
    
		\hspace{4em}[149] image `Magento Admin Panel'
        
	\hspace{2em}[150] navigation `'
    
		\hspace{4em}[151] menubar `', orientation=`horizontal'
        
			\hspace{6em}[153] link `\verb|\ue604| DASHBOARD'
            
				\hspace{8em}StaticText `\verb|\ue604|'
                
				\hspace{8em}StaticText `DASHBOARD'
                
			\hspace{6em}[156] link `\verb|\ue60b| SALES'

            \hspace{6em}...

\textbf{Failure Point:}

---------------------------------------------------------------------------------------------------

\textcolor{blue}{- Goal Decomposition:} Extract the search term with the highest `Uses' value from the `Top Search Terms' table and list it as the top search term.

\textcolor{blue}{\ \ - Corresponding Reasoning:}
\textcolor{blue}{<think>}

\textcolor{blue}{In the goal decomposition stage, the task answer has already been fully identified.}

\textcolor{red!70}{The first thing I notice is that the table under ``Top Search Terms'' already has entries.} For example, \textcolor{red!70}{``hollister'' has 19 uses,} ``Joust Bag'' has 4, ``Antonia Racer Tank'' has 2, ``tanks'' has 1, and ``WP10'' has 1. The user wants the top 1 search term, which would be the one with the \textcolor{red!70}{highest number in the ``Uses'' column.}\

…

Since the data is already present on the page, \textcolor{red!70}{the next logical step is to extract this information.}

\textcolor{blue}{</think>}

\textcolor{blue}{- Observation Decomposition:} See the region highlighted by the red rectangle in Figure~\ref{fig:qual_case_3_obs} for detail.  \textcolor{red!70}{\# Extract all necessary elements according sub-goal correctly from original observation.} 

\textcolor{blue}{Final Action:} click(``918'')  \textcolor{red!70}{\# [918] gridcell `hollister'}
\hspace{1em}
\raisebox{-0.4em}{\includegraphics[height=1.2em]{figs/cross.png}}

\textcolor{blue}{\ \ - Corresponding Reasoning:}
\textcolor{blue}{<think>}

\textcolor{blue}{1. The agent correctly analysis that `Hollister' is the finally answer.}

Looking at the provided data, the rows under `Top Search Terms' include entries like `hollister' with 19 uses, `Joust Bag' with 4, `Antonia Racer Tank' with 2, `tanks' with 1, and `WP10' with 1. \textcolor{red!70}{The highest `Uses' here is 19 for `hollister'. }

\textcolor{blue}{2. However, the agent is misdirected by the action ``extract'' in the sub-goal, which does not exist in the action space. As a result, the agent starts overthinking and attempts to find the closest action to ``extract.''}

But the assistant's role here is to perform actions on the webpage. Since the user is asking for the top search term, and the data is present in the table, \textcolor{red!70}{the assistant needs to extract that information.} However, \textcolor{red!70}{the available actions don't include reading text from elements. The actions available are things like click, fill, etc. }

\textcolor{blue}{</think>}

\end{casebox}

\subsection{Compwob+ Case Studies}
\label{app:case_compwob}
For static webpages such as CompWoB+, TTED can leverage successful trial steps as demonstrations for in-context learning (ICL), enabling it to directly generate final actions for a single task query. We provide a detailed description in Appendix~\ref{app:impl_TTED}.

We present a representative case from CompWoB+ to illustrate TTED in single-turn, static webpages. In this case, the task requires completing multiple sequential operations, including reading a table, clicking a download button, and selecting a radio option. TTED first decomposes the original goal into multiple sub-goals and extracts the corresponding sub-observations for each sub-goal.

Based on the decomposed sub-goal, the agent generates an initial sequence of actions. A JavaScript-based simulation is then applied to validate each action. The simulation reveals that clicking a table cell is unnecessary and ineffective, leading to rejection by the self-assessment module. The agent subsequently retries action generation, removes redundant interactions, and refines the action sequence. After validation through self-assessment process, the revised action sequence is accepted, leading to successful task completion.

This case demonstrates how environment decomposition simplifies compositional tasks, while the action generation – self-assessment loop further improves action reliability and robustness. 

\begin{figure}[h]
  \centering
  \setlength{\fboxsep}{0pt}
  \fbox{
  \includegraphics[width=0.8\linewidth]{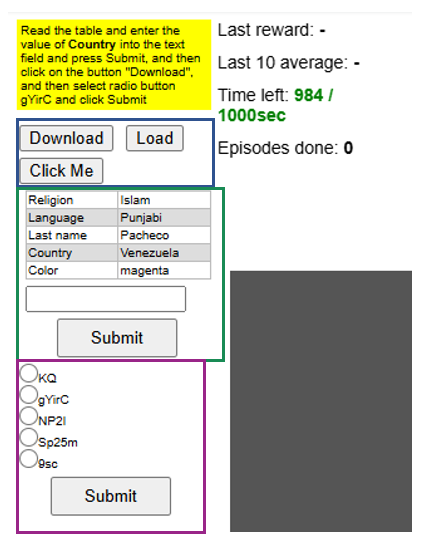}
  }
  \caption{Observation for \# Compwob+.}
  \label{fig:qual_case_compwob_obs}
\end{figure}

\begin{figure*}[t]
  \centering
  \includegraphics[width=0.95\linewidth]{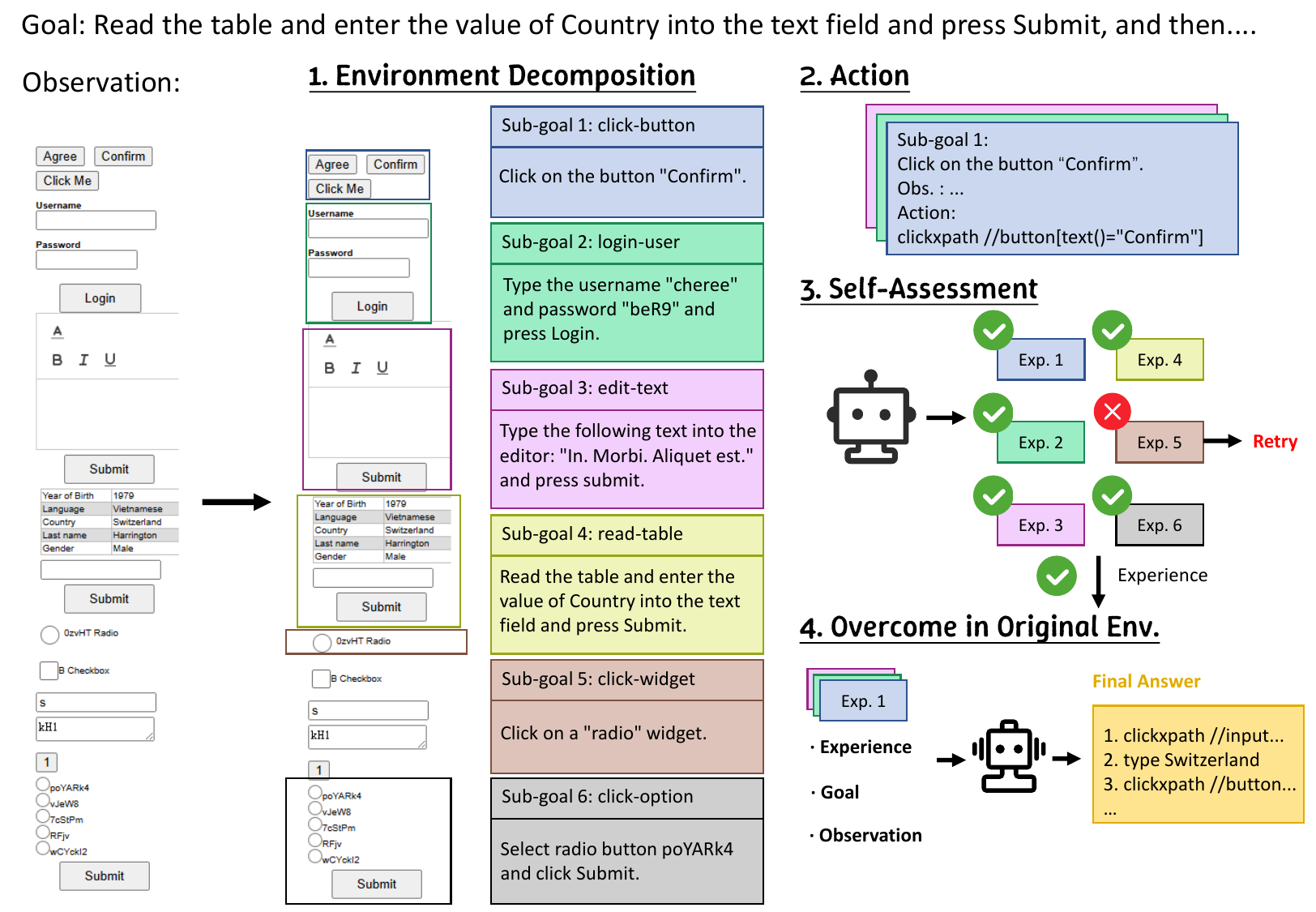}
  \caption{Overall workflow of TTED with in-context learning (ICL) on CompWoB+. Given a complex goal and the original observation, the agent first performs environment decomposition to divide the task into multiple sub-goals and extract the corresponding sub-observations. For each decomposed sub-environment, the agent generates trial actions and evaluates them through the self-assessment module. The successful trials are stored as experiences, while failed trials trigger retry until success or reaching the maximum retry limit. Finally, the collected successful experiences are leveraged as demonstrations for ICL, guiding the agent to directly generate the final action sequence in the original environment.}
  \label{fig:compwob_workflow}
\end{figure*}

\begin{casebox}{\# Compwob+ Case: Environment Decomposition}
\textbf{Goal:} Read the table and enter the value of Country into the text field and press Submit, and then click on the button ``Download'', and then select radio button gYirC and click Submit

\textbf{Page:} See Figure~\ref{fig:qual_case_compwob_obs}. for detail.

\textcolor{blue}{- Goal Decomposition: }

1. read-table: Read the table and enter the value of Country into the text field and press Submit  

2. click-button: Click on the button ``Download''  

3. click-option: Select radio button gYirC and click Submit

\textcolor{blue}{- Observation Decomposition: } See the region highlighted by rectangles in Figure~\ref{fig:qual_case_compwob_obs} for detail.  

\textcolor{blue}{\ \ - Part of Sub-Observation:}

subtask 2: click-button: Click on the button ``Download''. 

html:

<!DOCTYPE html>

<html>

\hspace{1em}<head>

      \hspace{2em}<title>Click Button</title>
      
\hspace{1em}</head>

\hspace{1em}<body>

      \hspace{2em}<div id=``buttonsContainer'' data-wob\_ref=``4'' data-wob\_eps=``e4''>
      
            \hspace{3em}<button class=``generated-button'' data-wob\_ref=``5'' data-wob\_eps=``e4''>Download</button>
            
      \hspace{2em}</div>
      
\hspace{1em}</body>

</html>

\end{casebox}

\begin{casebox}{\# Compwob+ Case: Trial Action Generation \& Simulation \& Self-Assessment}

\textcolor{blue}{- Sub-Goal:} Read-table: Read the table and enter the value of Country into the text field and press Submit.

\textcolor{blue}{- Sub-Observation:} See the region highlighted by green rectangle in Figure~\ref{fig:qual_case_compwob_obs} for detail. 

<!DOCTYPE html>

<html>

<head>

    \hspace{1em}<title>Read Table</title>
    
</head>

<body>

    \hspace{1em}<div id=``tab'' data-wob\_ref=``8'' data-wob\_eps=``e4''>
    
        \hspace{2em}<table data-wob\_ref=``9'' data-wob\_eps=``e4''>
        
            \hspace{3em}<tr data-wob\_ref=``10'' data-wob\_eps=``e4''>
            
                \hspace{4em}<td data-wob\_ref=``11'' data-wob\_eps=``e4''>Religion</td>
                
                \hspace{4em}...
                
\textcolor{blue}{- Action:}

1. clickxpath //tr[td[text()=`Country']]/td[2]

2. clickxpath //input[@id=`tt']

3. type Venezuela

4. clickxpath //button[@id=`tablesubbtn']

\textcolor{blue}{- JS Simulation:}

\textcolor{blue}{// Action 1. clickxpath //tr[td[text()=`Country']]/td[2]}

var countryCell = document.evaluate(``//tr[td[text()=`Country']]

/td[2]'', document, null, 

XPathResult.FIRST\_ORDERED\_NODE\_TYPE, null).singleNodeValue;

return countryCell === document.activeElement;

\textcolor{blue}{// Action 2. clickxpath //input[@id=`tt']}

var inputField = document.getElementById(``tt'');

return inputField === document.activeElement; // Verifies if the input field is focused after the click

\textcolor{blue}{// Action 3. type Venezuela}

var inputField = document.getElementById(``tt'');

return inputField.value === ``Venezuela'';

\textcolor{blue}{// Action 4. clickxpath //button[@id=`tablesubbtn']}

var submitButton = document.getElementById(``tablesubbtn'');

return submitButton === document.activeElement;

\textcolor{blue}{Simulation Result:} [False, True, True, True]

\textcolor{blue}{- Self-Assessment Result:} \textcolor{red!70}{Reject}

\textcolor{blue}{\ \ - Corresponding Reasoning:}
\textcolor{blue}{<think>}

1. Instruction 1: The XPath `//tr[td[text()=`Country']]/td[2]' correctly targets the ``Venezuela'' cell in the table. \textcolor{red!70}{However, the inspect result is **False**,} indicating the click action failed. This is likely \textcolor{red!70}{because clicking a table cell (which is not an input or button) does not interact with the content} (e.g., select text or copy it). Since the input field `\#tt' is already pre-filled with ``Venezuela'', this step is **unnecessary** and invalidates the task flow.

2. Instruction 2: Clicking `//input[@id=`tt']' is valid and focuses the input field (inspect result: **True**).

3. Instruction 3: Typing ``Venezuela'' into the input field is redundant since the field is already pre-filled. However, the observation confirms the value matches, so the action is technically valid (inspect result: **True**).

4. Instruction 4: Clicking the ``Submit'' button is valid (inspect result: **True**), though it has no functional effect as per the note.

\textcolor{blue}{</think>}

\textcolor{red!70}{\# Retry} Action Generation → Self-Assessment → Assessment result: \textcolor{red!70}{Pass}

\textcolor{blue}{Final Action:}

1. clickxpath //input[@id=`tt']

2. type Venezuela
\hspace{13em}
\raisebox{-0.4em}{\includegraphics[height=1.2em]{figs/right.png}}

3. clickxpath //button[@id=`tablesubbtn']

\end{casebox}

\begin{casebox}{Final Action Generation}

\textbf{Goal:} Read the table and enter the value of Country into the text field and press Submit, and then click on the button ``Download'', and then select radio button gYirC and click Submit.

\textbf{Page:} See Figure~\ref{fig:qual_case_compwob_obs} for detail.

\textbf{Observation:}

<div id=``wrap'' data-wob\_ref=``2'' data-wob\_eps=``e4''>

    \hspace{2em}<div id=``query''>
    
        \hspace{4em}Read the table and enter the value of ...
        
    \hspace{2em}</div>
    
    \hspace{2em}<div id=``area'' data-wob\_ref=``3'' data-wob\_eps=``e4''>
    
    \hspace{2em}...
    
\textbf{Experience:}

...

Sub-goal 2: Read-table: Read the table and enter the value of Country into the text field and press Submit.

Sub-observation 2: See the region highlighted by green rectangle in Figure~\ref{fig:qual_case_compwob_obs} for detail.

Action:

1. clickxpath //input[@id=`tt']

2. type Venezuela

3. clickxpath //button[@id=`tablesubbtn']

Sub-goal 3: ...

...

\textcolor{blue}{Final Action:}

1. clickxpath //input[@id=`tt']

2. type Venezuela

3. clickxpath //button[@id=`tablesubbtn']

4. clickxpath //button[text()=``Download''] 
\hspace{5em}
\raisebox{-0.4em}{\includegraphics[height=1.2em]{figs/right.png}}

5. clickxpath //input[@id=`ch1'] 

6. clickxpath //button[@id=`optionsubbtn']
\end{casebox}
\begin{figure*}[t]
  \centering
  \includegraphics[width=0.95\linewidth]{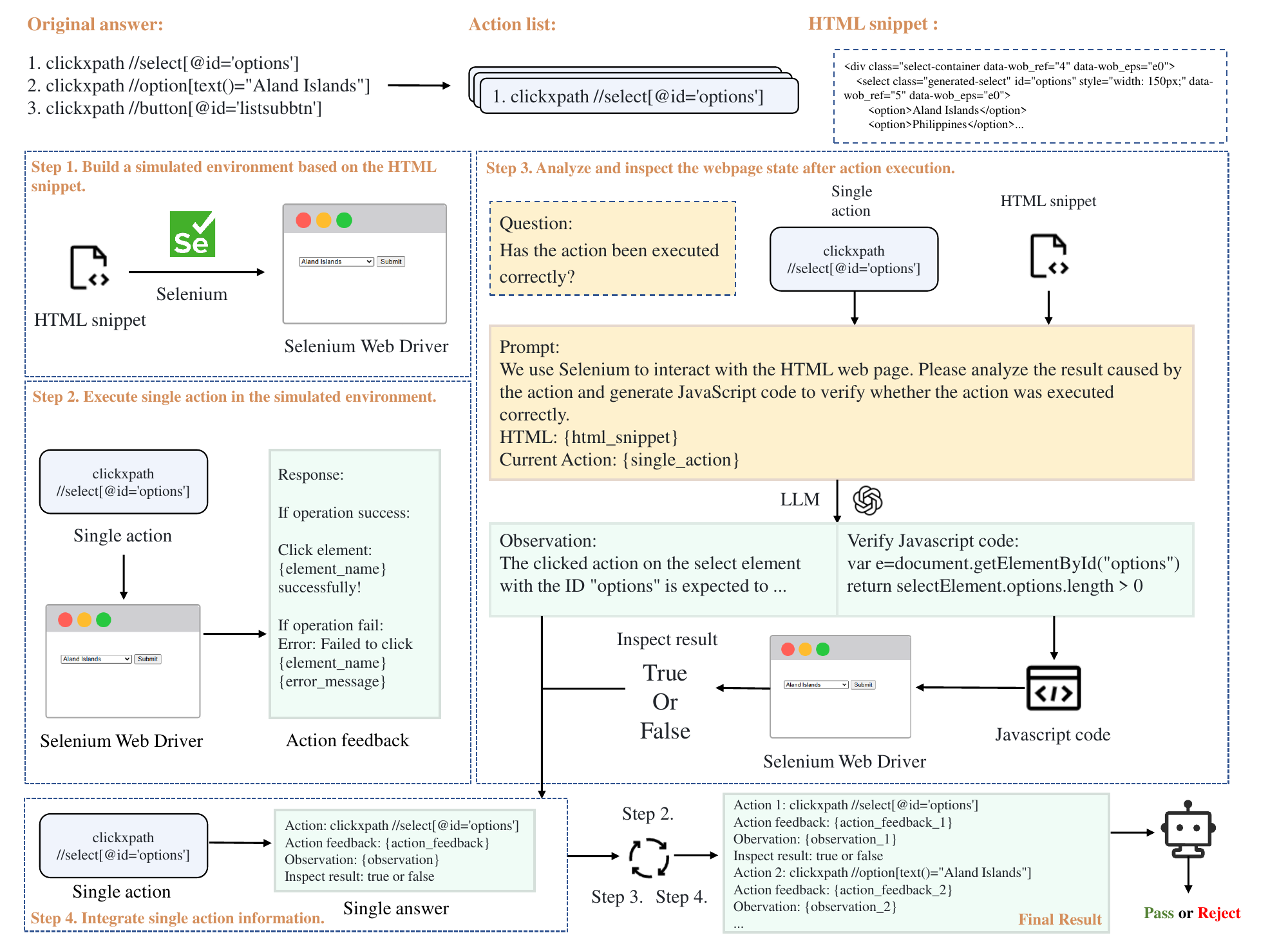}
  \caption{Workflow of the self-assessment module in TTED with ICL. For each generated action, a simulated environment is first constructed from the HTML snippet using Selenium. The action is then executed within the simulated environment, and the resulting webpage state is analyzed. A JavaScript verification script is generated to inspect whether the action is performed correctly. The execution feedback, observations, and verification results are integrated to determine whether the action is successful and returns the final judgment.}
  \label{fig:compwob_judge_workflow}
\end{figure*}

\section{Implementation Details}\label{app:imp-detail}

The models and datasets utilized in this study are intended strictly for research purposes. Furthermore, we have maintained full adherence to all applicable licensing agreements and established ethical frameworks.

\subsection{Implementation Details of Data Collection}
\label{app:data_collection}
In this section, we present dataset details, including the construction of CompWoB+~\citep{furuta2024exposing} and the subset of tasks selected from WebArena~\citep{zhou2024webarena}.

For CompWoB+, we design a task generator to address the limited number of high-composition tasks in the original dataset. The generator first selects multiple base miniwob~\citep{liu2018reinforcement} tasks according to a configuration file, which specifies both the task types and the composition size. These selected tasks are then composed into a unified web page to construct a composite environment. Each base task is implemented as an independent UI module, and for each module, the generator randomly samples a task instance. For example, a button with randomly generated text, a list with randomly sampled options, or a table with randomly generated entries may be created. All randomness is controlled by a global random seed. By varying the seed, the generator produces different task instances, enabling large-scale task generation. In summary, the generator supports arbitrary combinations of base tasks and can generate composite tasks with varying composition sizes.

Using this generator, we construct the CompWoB+ dataset. Specifically, we select 10 base tasks and use the generator to arbitrarily combine them into composite tasks. The list of base tasks is shown in Table~\ref{tab:compwob_base_task}. We generate composite tasks with composition sizes ranging from 3 to 8. For each composition size, we construct 10 distinct composite tasks, and for each task, we further generate 10 randomized instances for evaluation. The resulting dataset statistics are summarized in Table~\ref{tab:compwob_task_stats}. As a result, Compwob+ ensures consistency with the difficulty of the original dataset by selecting and combining unit modules from the original dataset, and task quantity is substantially enlarged (from 50 to 600).

\begin{table}[h]
  \centering
  \resizebox{0.7\linewidth}{!}{
  \begin{tabular}{llll}
    \toprule
    \textbf{ID} & \textbf{Base Task} & \textbf{ID} & \textbf{Base Task} \\
    \midrule
    1 & click\_button     & 6  & login\_user \\
    2 & click\_link       & 7  & choose\_list \\
    3 & click\_widget    & 8  & read\_table \\
    4 & click\_checkboxes & 9  & edit\_text \\
    5 & click\_option    & 10 & enter\_time \\
    \bottomrule
  \end{tabular}
  }
  \caption{Base tasks used for constructing composite tasks in CompWoB+.}
  \label{tab:compwob_base_task}
\end{table}

\begin{table}[h]
  \centering
  \resizebox{\linewidth}{!}{
  \begin{tabular}{cccc}
    \toprule
    \textbf{Composition Size} & \textbf{Task} & \textbf{Instances / Task} & \textbf{Total Instances} \\
    \midrule
    3 & 10 & 10 & 100 \\
    4 & 10 & 10 & 100 \\
    5 & 10 & 10 & 100 \\
    6 & 10 & 10 & 100 \\
    7 & 10 & 10 & 100 \\
    8 & 10 & 10 & 100 \\
    \midrule
    Total & 60 & -- & 600 \\
    \bottomrule
  \end{tabular}
  }
  \caption{Statistics of composite tasks in CompWoB+.}
  \label{tab:compwob_task_stats}
\end{table}

For WebArena, some tasks are difficult to collect and evaluate due to network-related issues. Therefore, we conduct experiments on a selected subset of tasks. The selected subset is shown in Table~\ref{tab:webarena_dataset}. Specifically, we use 561 of 812 WebArena tasks, and Reddit could not operate reliably under our network, whose completion rate is lower than 5\%. Even after this exclusion, our benchmark remains larger than the commonly used WebArena-Lite (165 tasks).

For text-based answers in WebArena, the evaluation relies on LLM-based fuzzy matching. In the original WebArena benchmark, GPT-4o-Turbo is used for answer evaluation. To improve the matching quality, we replace it with GPT-5-mini~\citep{singh2025openaigpt5card}. It is worth noting that this process is used only internally by the benchmark to determine whether a task is successfully completed, and is not used as a reward model to guide TTED.

\begin{table}[h]
  \centering
  \resizebox{\linewidth}{!}{
  \begin{tabular}{ccccccc}
    \toprule
    & CMS & Map & Shopping & Gitlab & Reddit* & Overall \\
    \midrule
    \#Tasks & 149 & 124 & 174 & 114 & 0 & 561 \\
    \bottomrule
  \end{tabular}
  }
  \caption{Statistics of the selected WebArena task subset. *For Reddit, we remove all tasks due to network issues.}
  \label{tab:webarena_dataset}
\end{table}

\subsection{Implementation Details of Model Training}\label{app:training_details}

In this section, we provide the hyper-parameter settings and implementation details for test-time training of the TTED framework, as well as the baseline methods discussed in Experiments section.

\paragraph{Infrastructure and Base Model}

The training experiments are conducted using the open-source Qwen3-8B in main experiments and Qwen3-14B~\citep{yang2025qwen3technicalreport} in additional experiments as the base model. We utilize the VeRL~\citep{10.1145/3689031.3696075} framework with VLLM~\citep{10.1145/3600006.3613165} inference engine to distribute the training workload efficiently. The experiments are executed on a cluster of 8 NVIDIA A100-SXM4-80GB GPUs. During the trial steps of TTED and data collection of baseline methods (as described in Methodology section), the sampling temperature is set to $1.0$, and thus obtain the rollout probabilities $\pi_{\theta_{\text{old,inf}}}$ directly from the inference engine. The training is performed on all parameters of the model. For RL, the maximum input token length is restricted to 16,384 tokens, and the maximum output token length is set to 4,096 for action generation and 6,144 for environment decomposition; and for SFT, the whole trajectory is left-truncated to 22,528 tokens. The overall training time for all methods is shown in Table~\ref{tab:training-times}. 
In addition to total GPU hours, Table~\ref{tab:training-times} also reports the test-time exploration costs, including average inference time, token usage, and interaction steps per sub-environment exploration. TTED incurs higher per-step exploration costs than TTT, but its total training cost remains comparable to TTT and lower than TTRL. Together with the performance improvement on WebArena (21.0\% vs.\ 12.5\%), these results show that TTED trades additional exploration costs for more effective adaptation rather than being cost-free. Table~\ref{tab:token} further shows that TTED reduces prompt lengths for action generation. Training with ground-truth labels requires less time because the overall accepted training data is fewer.

\begin{table}[h]
\centering
\resizebox{\linewidth}{!}{
\begin{tabular}{llcccc}
\toprule
\textbf{Adv. Estimation} & \textbf{Method} & \textbf{Inf. Time (s)} & \textbf{\#Tokens} & \textbf{\#Steps} & \textbf{GPU h.} \\
\midrule
\multirow{3}{*}{SFT}
& SFT w/ GT & -- & -- & -- & 2.24 \\
& TTT & \phantom -- & -- & -- & 11.07 \\
& TTED & \phantom -- & -- & -- & 12.53 \\
\midrule
\multirow{3}{*}{REINFORCE}
& RL w/ GT & -- & -- & -- & 8.67 \\
& TTT & \phantom{0}16.7 & 17514.9 & 6.17 & 14.40 \\
& \textbf{TTED} & \textbf{\phantom{0}46.8} & \textbf{26114.5} & \textbf{9.91} & \textbf{14.80} \\
\midrule
\multirow{2}{*}{GRPO}
& TTRL & \phantom{0}34.9 & 25808.0 & 4.54 & 19.20 \\
& TTED & 101.9 & 29231.7 & 5.76 & 21.33 \\
\bottomrule
\end{tabular}
}
\caption{Computational costs for different adaptation methods on WebArena. All methods are evaluated under a uniform training budget of 3200 sub-environments. GPU hours are estimated with Qwen3-8B on 8 NVIDIA A100-SXM4-80GB GPUs. Avg. inference time and token numbers are measured per sub-environment exploration.}
\label{tab:training-times}
\vspace{-0.5em}
\end{table}

\paragraph{Data Collection and Trajectory Rollout} 
For the trial steps defined in Methodology section, the agent interacts with the decomposed sub-environments of WebArena to gather trajectories, including accepted and rejected trajectories of action generation and environment decomposition. In addition, we filter out action generation trajectories with actions outside the action space with rules. 
For all methods, the training batch is constructed by sampling trajectories from 32 sub-environments, with one action generation and two environment decomposition trajectories, i.e. sub-goal and sub-observation decomposition (if applicable), per sub-environment. To ensure a fair comparison between SFT and RL methods, we obtain all data before training and reuse the same set of trajectories for both SFT and RL training. For methods with GRPO, we collect an additional set of trajectories by sampling multiple actions per sub-environment, resulting in the total group size of four, to compute the self-consistency reward. For the self-consistency matching, two actions are considered a match if they are identical in their action type and parameters, except (1) the ``noop'' and ``report\_infeasible'' actions are ignored for parameter matching; and (2) the ``fill'' and ``send\_msg\_to\_user'' actions are considered a match when their textual parameters are semantically similar, or concretely, over 0.75 cosine similarity of embeddings on all-MiniLM-L6-v2~\citep{reimers-gurevych-2019-sentence,10.5555/3495724.3496209}. And the majority will be assigned rewards equal to one, and others are rewarded zero. 
The training process is conducted for at maximum 100 steps, resulting in a total of 3200 sub-environments for training. 

\paragraph{Model Training in TTED}
As detailed in Methodology section, we adopt an off-policy REINFORCE algorithm augmented with rollout importance sampling and advantage normalization from REINFORCE++~\citep{hu2025reinforcestabilizingcriticfreepolicy}. The policy optimization employs the Group Sequence Policy Optimization (GSPO) loss~\citep{zheng2025groupsequencepolicyoptimization}.
\begin{itemize}
    \item \textbf{Reward Normalization:} Rewards are assessed pointwise and independently for each trajectory by using the backbone model as a GRM. The according prompt template, including principles and scoring pattern, is elaborated in Appendix~\ref{app:prompt}. The raw scores from the GRM---specifically $r_{\text{goal}}, r_{\text{obs}} \in [0,5]$ and $r_{\text{action}} \in \{0,1\}$---are mapped to a bounded scalar return $r^*_t \in [-1, 1]$ with linear transformation. 
    \item \textbf{Advantage Estimation:} For the default REINFORCE-based setting, we compute the normalized advantage $\hat{A}_t = \frac{r^*_{t}-\mu_{r^*}}{\sigma_{r^*}+\epsilon}$ across the training mini-batch to stabilize the test-time training process. For the SFT variant, we remain trajectories with rewards larger than 0 after normalization; for the GPRO variant, we use self-consistency rewards as described above, apply the original GRPO advantage estimation, and filter out groups whose actions receive all the same reward value. 
    \item \textbf{Optimization Hyper-parameters:} The peak learning rate for RL is set to $10^{-6}$ with none-warmup schedule and a 0.1 decay; and the learning rate for SFT is $10^{-5}$. These setings are resulted from a grid search on $\{5\times10^{-7}, 10^{-6}, 2\times10^{-6}, 5\times10^{-6}, 10^{-5}\}$. To prevent model degradation and adaptation drift, we apply a gradient clipping with a threshold of 5.0 and, for RL, IS ratio clipping with an upperbound of 0.01 and a lowerbound of 0.005. We do not use KL divergence regularization, as we find it does not contribute to the performance improvement in our preliminary experiments. 
\end{itemize}
Since it is not allowed for us to test the checkpoints on WebArena in the middle of test-time training, 
we also split an 0.1 ratio validation set, and observe the GSPO loss value to stably descend on the validation set as the training goes. So that, the final checkpoint is selected as the last checkpoint. 

\paragraph{Model Training in Baseline Methods}
For the ``SFT w/ GT'' and test-time SFT (``TTT (SFT)'') baselines evaluated in Table~1, the models are fine-tuned purely on the successful demonstration trajectories decided by the ground truth labels or agent self-assessed rewards (larger than 0 after normalization), respectively. For ground truth labeling, if a multi-turn trajectory is labelled success, we treat its every single turn as a success for training. Even though, the training set of ``SFT w/ GT'' is too small that we train the model with two epoches (still within the maximum of 100 training steps). For test-time RL (``TTT (RL)'') and TTRL, we also use GSPO loss with REINFORCE++-based and GRPO-based advantage estimation approaches, respectively. Other settings are the same to the corresponding variants of TTED. 

\subsection{Implementation Details of TTED}\label{app:impl_TTED}

In this section, we provide detailed settings for TTED, including hyperparameters and model configurations for each dataset evaluation.

\paragraph{Evaluation on CompWob+.} We evaluate TTED with in-context learning (ICL) on CompWoB+~\citep{furuta2024exposing}. We describe the experimental setup and hyperparameter settings in the following.

\subparagraph{Base Model:} For evaluation on CompWob+, we use GPT-4o-mini-2024-07-18~\citep{openai2024gpt4ocard} and Qwen3-8B~\citep{yang2025qwen3technicalreport} as the base model. We access GPT-4o-mini-2024-07-18 via the API. For Qwen3-8B, we deploy the model using VLLM~\citep{10.1145/3600006.3613165} and run inference on 4 NVIDIA A100-SXM4-80GB GPUs.

\subparagraph{Pipeline:} Since CompWoB+ consists of single-turn static webpages, we adopt TTED with in-context learning (ICL) for evaluation. Unlike real-world web environments, the entire environment in CompWoB+ can be decomposed at once into multiple sub-environments. The agent first attempts to solve tasks within these sub-environments, and the resulting trajectories are used as experience to guide final action generation in the original environment via ICL. The overall workflow is illustrated in Figure~\ref{fig:compwob_workflow}. In particular, for the self-assessment process in the sub-environment, we further improve reliability through simulation. Specifically, we first construct a simulated environment using Selenium~\citep{selenium} based on each sub-environment. The generated actions are then executed in the simulated environment to obtain execution feedback. In addition, we employ the LLM to generate JavaScript code to verify whether the actions are correctly executed. Finally, the execution feedback, JavaScript-based verification signals, and generated actions are provided to evaluate whether the action in the current sub-environment is correct. If the actions is judged as successful, the attempt in the sub-environment is recorded as experience and used to guide the final action generation in the original environment. Otherwise, the agent retries within the sub-environment. See Figure~\ref{fig:compwob_judge_workflow} for detail.

\subparagraph{Hyper-parameters:} 
We introduce several hyperparameters to balance reliable pipeline execution and efficiency. 
When errors occur during execution, we employ retry mechanisms to improve robustness. 
Specifically, if the LLM output does not satisfy the required format and answer extraction fails, the model retries up to 5 times. 
Additionally, when the self-assessment module determines that the actions fails within the sub-environment, the agent re-attempts completion within the same sub-environment for up to 3 attempts.
If the actions still fails after three attempts, the current sub-environment is discarded. 
For each interaction with the LLM, we limit the maximum input length to 30,000 tokens, output length to 5,000 tokens. 
For Qwen3-series models, we further allocate an additional thinking budget of 5,000 tokens.

\paragraph{Rollouts on WebArena.} To demonstrate the ability to tackle realistic web automation tasks, we adopt TTED with RL to improve performance on WebArena. The rollout strategy and settings are described as follows.

\subparagraph{Base Model:} We use Qwen3-8B~\citep{yang2025qwen3technicalreport} as the base model for TTED in the WebArena sampling. We deploy Qwen3-8B using the VLLM~\citep{10.1145/3600006.3613165} inference framework and run inference on 4 NVIDIA A100-SXM4-80GB GPUs.

\subparagraph{Pipeline:} The rollout pipeline is shown in Figure~3.
At each step, the agent performs environment decomposition based on the current environment, producing a sub-goal and a corresponding sub-observation. Conditioned on the decomposed sub-environment, the agent then generates an action for the current step. Before executing this action in the original environment, the self-assessment module evaluates both the quality of the decomposition and the correctness of the generated action, including whether the sub-goal and sub-observation are appropriate for the current state. Only when both the decomposition and the generated action pass self-assessment is the action executed in the original environment to advance the task. Otherwise, the agent re-performs environment decomposition on the current environment and starts a new attempt. The summary module then records the action information as part of the interaction history.

Through this process, we collect a large number of sub-environments along with their corresponding self-assessment signals on WebArena. These sampled trajectories are then used as training data for subsequent RL optimization.

\subparagraph{Hyper-parameters:} During rollout on WebArena, we set the maximum number of environment steps to 15. If the step limit is exceeded, the task is terminated. When the LLM output fails to satisfy the required format, we retry generation for up to five times. If the output still fails after 5 attempts, the task is considered failed and terminated. We set the sampling temperature to 1.0. In the self-assessment stage, the action is evaluated as either pass or reject, while the sub-goal and sub-observation are scored on a scale from 0 to 5. We provide a detailed description of the evaluation principle for actions, sub-goals, and sub-observations in Appendix~\ref{app:prompt}. A sub-environment is considered successful only when the action is evaluated as pass and both the sub-goal and sub-observation receive scores of at least 3. For each LLM call, the maximum input length is limited to 30,000 tokens, and the maximum output length is limited to 5,000 tokens. In addition, we allocate a reasoning token budget of 5,000 tokens. If any of these limits are exceeded, the corresponding inputs or outputs are truncated.

\paragraph{Evaluation on WebArena.} We evaluate TTED with RL on WebArena~\citep{zhou2024webarena}, a real-world web environment benchmark. The experimental setup and hyperparameter settings are described as follows.

\subparagraph{Base Model:}
We use Qwen3-8B~\citep{yang2025qwen3technicalreport} and our trained versions as the base model for TTED in the WebArena evaluation. We deploy Qwen3-8B using the VLLM~\citep{10.1145/3600006.3613165} inference framework and run inference on 4 NVIDIA A100-SXM4-80GB GPUs.

\subparagraph{Pipeline:} During evaluation, we remove the self-assessment module, while keeping the rest of the pipeline consistent with the sampling stage. Specifically, the agent first performs environment decomposition to obtain sub-environments, then generates actions within each sub-environment, and finally records the actions through the summary module. This process is repeated iteratively until the task is completed or reach the maximum steps.

\subparagraph{Hyper-parameters:} During evaluation, the maximum number of environment steps is set to 15. For the LLM, we set the temperature to 0. All other settings follow those used in the rollout stage, including a maximum input length of 30,000 tokens, a maximum output length of 5,000 tokens, and a reasoning token limit of 5,000. When the LLM output fails to satisfy the required format, we retry generation for up to 5 attempts.

\section{Prompt Templates}\label{app:prompt}

In this section, we provide the LLM prompts used in the implementation of TTED, including the TTED with in-context learning (ICL) method applied to single-page web tasks in CompWoB+~\citep{furuta2024exposing}, and the TTED with RL method used in realistic web environments WebArena~\citep{zhou2024webarena}.

\subsection{Prompt for Single-Page Web Tasks}
\label{app:compwob_prompt}
We perform environment decomposition through two stages: Goal Decomposition and Observation Decomposition. 
First, we decompose the overall task goal into multiple sub-goals. In our implementation, we provide the names of the base tasks that constitute the CompWoB+ dataset, along with the current task name. Based on this information, the model extracts the corresponding sub-goals from the overall task objective. 

Next, in the Observation Decomposition stage, the model extracts the corresponding sub-observations based on the identified sub-goals and the full observation. Each pair of sub-goal and sub-observation forms a sub-environment.

\begin{casebox}{Goal Decomposition}

Given the composite task name and its goal description, decompose the composite goal into exactly some sub-goals. For each sub-goal, output the sub-goal name, and a corresponding concise natural language instruction. There are ten types of sub-goals: `login-user', `click-link', `click-widget', `click-button', `click-option', `enter-time', `choose-list', `click-checkboxes', `edit-text', `read-table'.
Note: A sub-goal does not imply a particular operation or goal, one sub-goal may involve multiple operations, for example, `click submit button' is often combined with other operations to form a single sub-goal.

\# Current Task

Task Name: \{task\_name\}

Goal: \{goal\}

Decomposed Sub-goals:

\end{casebox}
\begin{casebox}{Observation Decomposition}

\# Instruction

Given the basic information of a compositional goal, the corresponding original web observation (HTML), and a list of its decomposed sub-goals, your work is to decompose the original complex environment into a series of simpler sub-environments, each includes a specific sub-goal and a corresponding sub-observation.

For each sub-observation:

Requirement 1: Preserve DOM Consistency with the Original Task

Each sub-observation must retain the same DOM structure and attributes (e.g., id, class, name, type, value, placeholder, etc.) of the target element as in the original HTML. This ensures that the same Selenium code (XPath, CSS selectors, etc.) used in the original observation will work identically in the decomposed observations.

Requirement 2: Include Required Context and Functional Components

In addition to the target element, each sub-observation must include any necessary surrounding or functional components required for the element to be visible, interpretable, and operable. These may include:
<form> tags and their attributes like action, method,
Related labels, container <div>s, submit buttons,
Necessary parent elements, classes, or structure used in the original HTML for styling or scripting purposes.

Requirement 3: Ensure Standalone Executability

Each sub-observation should be a self-contained and standalone HTML file with a valid document structure (<!DOCTYPE html>, <html>, <head>, <body>, etc.). It must be directly openable in a browser and fully functional for manual interaction and Selenium-based automation without relying on external scripts or styles from the original page.

Requirement 4: Ensure Output Confirms to the Required Format

The name of each decomposed sub-goal in your answer should be wrapped in <subgoal> sub-goal content </subgoal>, and HTML content should be wrapped in <subobs> html content </subobs>.

Example:

<subgoal> subgoal 1: ... </subgoal>

<subobs>

 HTML Content
 
</subobs>

\# Current tasks:

Task Name: \{task\_name\}

Goal Description: \{goal\}

Decomposed Sub-goals: \{sub\_goals\}

Original HTML: \{ori\_html\}

Decomposed HTML:
\end{casebox}

\begin{casebox}{Trial Action Generation}
You are an autonomous computer control agent that can perform atomic instructions specified by natural language to control computers. There are two types of instructions it can execute. 

\{action\_space\_description\}

\# Current Task

Below is the HTML code of the webpage where the agent should solve a goal.

Current observation: \{sub\_obs\}

Current goal: \{sub\_goal\}

Directly output the instructions, and do not include any thought and explanatory comments in your response.

Answer:

\end{casebox}

For the Script-Based Simulation component, we first predict the expected outcome of the generated action, and then generate corresponding JavaScript code to verify the predicted effect. Additionally, we explicitly inform the model that, in CompWoB, the submit button does not behave as a typical submission action and does not trigger page navigation.

\begin{casebox}{Script-Based Simulation}
You are given a HTML snippet, a goal and a current user action that is executed on this HTML using Selenium.

Your tasks are:

1. Analyze what change or effect the action is expected to cause in the web page.

Attention! All buttons named ``submit'' or containing ``submit'' do not have actual submission functionality; clicking them does not trigger any effect. Therefore, to verify this action, it is sufficient to check whether the button is being clicked.

2. Generate JavaScript code that can be executed to verify whether the action was correctly performed. The generated JavaScript code must include a return value, and the return value must be either true or false. Do not wrap the generated JavaScript code in a function. Note! The generated JS code must comply with JavaScript syntax standards and ensure that the code can be executed properly.

\# Example JavaScript code

var selectElement = document.getElementById(``options'');

return selectElement.value === ``Cameroon'';

3. Return the result in the following format:

\# Response Format:

Analysis:

<text>

<Brief explanation of the expected change on the page>

</text>

Verify Javascript code:

<code>

<JavaScript code that checks whether the action was successful>

</code>

\# Example

Action: clickxpath //select[@class=`generated-select' and @id=`options']

Analysis:

<text>

Observation: The action is expected to trigger the dropdown selection process, allowing the user to choose one of the options displayed in the `<select>' element. However, since the action performed is just clicking on the `<select>' element, it might not directly result in a change until a specific option is selected. The verification can check if any interaction was triggered successfully.

</text>

Verify Javascript code:

<code>

var selectElement = document.getElementById(``options'');

return selectElement === document.activeElement; // Verifies if the select element is focused after the click

</code>

\# Current task:

Goal: \{sub\_goal\} 

HTML: \{sub\_obs\}

Action: \{action\}

JS code:

\end{casebox}

For the Self-Assessment stage, the model evaluates whether the generated action is successful by considering the observation, the sub-environment, and the execution results obtained from the simulated environment. Based on this information, the model provides a final judgment of either pass or reject.

\begin{casebox}{Self-Assessment}
You are a helpful assistant. What you need to do is determine whether the instruction can be successfully executed on the web page and whether it achieves the corresponding goal.

\{action\_space\_description\}

Current HTML Page: \{sub\_obs\}

Current Goal: \{sub\_goal\}

Current Instructions: \{actions\}

Every instructions is executed on the web page and corresponding feedback can be obtained. Specifically, for each command, the following information is provided:

Action: The action that was executed

Observation: The change observed on the web page after the action

Inspect result: Whether the action was executed correctly

Here are the instructions' information:

\{js\_simulation\}

Your Job: 

1. Analyze: Determine whether the given instructions can successfully accomplish the goal on the provided HTML page. Please analyze each instruction one by one to determine whether an error occurred.

2. These instructions are executed within the web page; please make judgments based on the feedback from their execution.

3. Attention! All buttons named ``login'', ``reset'', ``submit'' or containing ``submit'' do not have actual submission functionality; clicking them does not trigger any effect. Similarly, descriptions in the goal such as click ``submit'', ``login'' and ``reset'' only require clicking the corresponding buttons, without the need for actual submission functionality.

4. Give your judge result. Attention output the final answer enclosed in \$\$. 

If they are correct, respond with: \$\$pass\$\$.

If they are incorrect or incomplete, respond with: \$\$reject\$\$

5. Based on the analysis, provide the recommend answer to the instruction. If the evaluation result is ``pass'' then the recommend answer is the original answer; if the result is ``fail'' generate a new recommend answer based on the analysis.

Response Format:

Analyze:

1. ...\# Instruction 1 analysis

2. ...\# Instruction 2 analysis

...

Judge result: \$\$pass\$\$ or \$\$reject\$\$

Recommend answer:...
\end{casebox}

\begin{casebox}{Final Action Generation}
You are an autonomous computer control agent that can perform atomic instructions specified by natural language to control computers. 

\{action\_space\_description\}

Here are some examples for specific tasks, please follow the example to generate your answer for the current task.

\{sub\_environments\}

Below is the HTML code of the webpage where the agent should solve a goal.

\# Current task

Current html: \{observation\}

Current goal: \{goal\}

Answer:
\end{casebox}

\subsection{Prompt for Realistic Web Automation Tasks}
\label{app:webarena_prompt}

For realistic web automation tasks, we also perform goal decomposition followed by observation decomposition. Unlike single-page web tasks, it is not feasible to generate all sub-environments at once. Instead, the process is performed iteratively. 
At each step, the agent generates the next sub-goal based on the overall objective, the current observation, and the historical trajectory. The observation decomposition module then extracts the relevant sub-observation according to the generated sub-goal and the current observation, thereby forming a sub-environment.

\begin{casebox}{Goal Decomposition}
You are a web agent planner.

Your job is to analyze the current web environment and the agent's progress so far, and then determine the most appropriate next sub-task the agent should perform.

You are given the following information:

1. **Task Goal**: The ultimate objective the user wants to achieve.

2. **Current Page (AXTree)**: A structured textual representation of the current page's accessible elements.

3. **Action History**: The list of actions the agent has performed so far in order.

4. **Failure Attempts**: A list of previously attempted actions that failed on this page, along with the reasons for their failure.

Your job:

- Reflect on the task goal, the current state of the page, and what has already been tried.

- Identify the most reasonable and necessary next sub-task to move the agent closer to the goal.

- The sub-task should be high-level but specific, and only describe the next immediate step.

     - Do not provide low-level commands such as click(`277'). Instead, describe the sub-task using high-level natural language — for example: ``Click the link labeled `The A11Y Project / a11yproject.com' to open the project page.''
    
- The sub-task should be as simple as possible, each containing only one objective and requiring only a single step to complete.

**Output format:**

\#\#\# Next Sub-task:

[One concise, natural language instruction describing the agent's next step.]

\# Current task

Task Goal: \{goal\}

Current Page (AXTree):

\{axtree\}

Action History:

\{history\}

Failure Attempts:

\{history.get\_error\_descendants()\}

\end{casebox}

\begin{casebox}{Observation Decomposition}
You are an intelligent web agent. Your task is to extract the goal-relevant sub-observation from a full **Accessibility Tree (AX Tree)** of a webpage.

Given:

- A user goal description.

- The full AX Tree text representation of the current web page.

Instructions:

- Select all AX Tree fragments (i.e., nodes and their subtrees) that are directly or indirectly relevant to the goal or sub-goal.

- Err on the side of inclusion: **if you are unsure whether an element is needed, include it.**

- You must **copy the selected fragments exactly as-is** from the original AX Tree.

- **Do not modify** any content — including the AXNode ID, node type, labels, values, or attributes.

- AXNode IDs (e.g., `[163]') must **match the original AX Tree exactly**.

- You may extract multiple non-contiguous branches if needed.

- Capture as much of the environment as possible to avoid omissions or errors.

---

\#\#\# Output format:

[The extracted sub-observation]

\# Current task:
Goal:
\{sub\_goal\}

\# Current Page

Accessibility Tree:

\{axtree\}

\end{casebox}

\begin{casebox}{Action Generation}
You are a web assistant. You will be given web-based tasks. These tasks will be accomplished through the use of specific actions you can issue.

Here's the information you'll have:

The user's objective: This is the final goal you're trying to achieve.

The sub-goal: This is the current sub-goal you need to perform in order to move toward the final goal.

The current web page's accessibility tree: This is a simplified representation of the webpage, providing key information.

The current web page's URL: This is the page you're currently navigating.

The open tabs: These are the tabs you have open.

The history actions: These are the history actions you have done, including actions and its intent.

The failure attempts: These are failed attempts made on the current page, including a failure summary, cause analysis, and suggestions. Please avoid making similar mistakes when performing actions.

You can interact with the environment using the following actions:

\{action\_space\_description\}

To be successful, it is very important to follow the following rules:

1. You should only issue an action that is valid given the current observation

2. You should only issue one action at a time.

Your output should be included:

- Clearly explain what the action is trying to achieve (intent)

- Clearly state which element is being operated on (element ID + label/type)

- Output the final action command that should be executed

Output format:

\#\#\# Manipulate element:

[Identify the specific element that the agent will interact with.]

\#\#\# Action:

[Write the exact action command that should be executed on the page.]

\#\#\# Action intent:

[Summarize what the action is trying to accomplish in the context of the task.]

Here is an example of how to use the bid action:

click(`314')

Here is an example:

\# Example

\{simple\_example\}

\# Current Task

Observation:

\{sub\_obs\}

URL: \{obs[``url'']\}

Final goal: \{obs[``goal'']\}

Sub-goal: \{sub\_goal\}

History Actions:

\{history\}

Failure attempts:

\{history.get\_error\_descendants()\}

Action:
\end{casebox}

In the self-assessment stage, we evaluate whether the generated action and the corresponding sub-environment are correct. To define correctness, we specify a set of principles that guide model for accurate evaluation. For actions, the generated action must be executable and capable of achieving the sub-goal. The action is evaluated in a binary manner, with outcomes labeled as either pass or reject.

For the sub-goal and sub-observation, we design a scoring scheme ranging from 0 to 5. Specifically, we expect the sub-goal to be closely related to the overall objective, achievable with a single action (i.e., atomic), clearly defined without ambiguity, and free from hallucinations. For the sub-observation, we require that the elements remain consistent with the original observation, include all elements necessary to accomplish the sub-goal, and preserve the structural organization of the original observation. Based on these criteria, the final scores are assigned.

\begin{casebox}{Self-Assessment}
You are an expert web automation evaluator, responsible for assessing both action execution success and the quality of goal/observation decomposition made by an AI agent during a web task.

Your judge combines two perspectives:

Action Execution Accuracy — Did the executed action actually achieve its intended purpose?

Decomposition Quality — Was the sub-goal (and its supporting sub-observation) well-defined, logically placed, and consistent with the overall task?

You will be provided with:

1. Overall Goal — the agent's high-level objective.

2. Overall Observation of Current Step — the full observation before sub-observation extraction.

3. Action History — a list of previously executed steps.

4. Executed Action — the specific action the agent just performed.

5. Detailed Element — the DOM element involved in the action.

6. Action Intent — the natural-language description of what the agent wanted to achieve with the action.

7. Sub-goal — the decomposed step corresponding to this action.

8. Sub-observation — the extracted sub-observation supporting that sub-goal.

Part 1: Action Execution Success

You must determine whether the executed action successfully fulfilled its intent.

- Based on the action and observation, determine whether the action can complete the sub-goal.

- If the action did not align with sub-goal exactly but still contributes positively toward achieving the overall goal, count it as successful.

- Provide a brief explanation of why the action succeeded or failed.

Output either:

true or false inside code fences for the result.

Part 2: Goal and Observation Decomposition Quality

Sub-goal Quality (0-5)

A high-quality sub-task should be atomic, precise, and logically aligned with both the task goal and previous steps.

Scoring reference:

\{goal\_decomposition\_principle\}

Here are some examples for goal quality judging:

\{sub\_goal\_judge\_example\}

Sub-observation Quality (0-5)

A valid sub-observation must be a verbatim subset of the original observation and contain all elements necessary to perform the sub-goal.

Scoring reference:

\{obs\_decomposition\_principle\}

Required Output Format (JSON)

\{

  ``action\_success'': true or false,
  
  ``action\_reasoning'': ``why the action succeeded or failed'',
  
  ``subgoal\_score'': int (0-5),
  
  ``subobs\_score'': int (0-5),
  
  ``subgoal\_reasoning'': ``1-2 sentences explaining the sub-task quality judge'',
  
  ``subobs\_reasoning'': ``1-2 sentences explaining the sub-environment quality judge'',
  
\}

\# Current Evaluate

Overall Task Goal:
\{goal\}

Overall Environment of Current Step:
\{obs\}

Previous Steps:
\{history\}

Current Step Sub-task:
\{subgoal\}

Current Step Sub-environment:
\{subobs\}

Executed Action:
\{action\}

Detailed Element:
\{element\}

Action Intent:
\{action\_intent\}
\end{casebox}

\begin{casebox}{Self-Assessment Principles}
\label{tab:priciples}
\#\#\# Goal Decomposition Principle:

5: Clear, atomic, non-redundant, consistent with the goal.

4: Slightly vague but mostly correct.

3: Somewhat unclear or contains mild future assumptions.

2: Poorly scoped or confusing.

1: Directionally wrong or ambiguous.

0: Not actionable or irrelevant.

\#\#\# Observation Decomposition Principle:

5: Fully correct, complete, and structurally consistent.

4: Minor omissions but still workable.

3: Noticeable omissions yet partially usable.

2: Missing key elements, difficult to use.

1: Nearly unusable.

0: Contains fabricated or altered elements.

\end{casebox}

\begin{casebox}{Summary}
You are a Summarization Agent in a web automation system.

Your role is to produce a concise, accurate, and context-aware summary of the current step.

Your summary must reflect (1) the current page environment, (2) what element was manipulated, (3) the action’s intent, and also incorporate the judge result to reflect whether the step was meaningful or successful.

You are given:

1. **Task Goal**: The overall objective of the task.

2. **Overall observation of Current Step**: The full environment of current step.

3. **Sub-goal**: The current sub-goal the agent was performing.

4. **Action**: The action will be excuted.

5. **Manipulated Element**: The element that is manipulated by the action.

6. **Action Intent**: A description of what the action is trying to achieve.

7. **Judge Result**: The evaluation result indicating whether the action succeeded in achieving its intent and the quality of the sub-task and sub-environment.

Your summary must include:

1. **Page Observation Description**: Briefly describe the functional nature of the current page, such as:

   - ``an online shopping page with a product search bar''
   
   - ``a login page requiring username and password input''
   
Do NOT list specific DOM nodes. Summarize the observation as a key functional areas relevant to the action.

2. **Manipulated Element**: Describe which element was interacted with, in natural language.

3. **The Summary of the Action**: Explain what the action achieve based on the given information.

Do not include reasoning or explanation — just the summary.

Required Output Format (JSON)

\{

    ``observation\_description'': ``a brief description of the functional areas of the page'',
    
    ``manipulated\_element'': ``the element that is manipulated by the action'',
    
    ``action\_summary'': ``a description of what the action achieve'',
    
\}

\# Current task

Task Goal: \{goal\}

Overall observation of Current Step: \{axtree\}

Sub-goal: \{subgoal\}

Action: \{action\}

Manipulated Element: \{element\}

Action Intent: \{intent\}

Judge Result: \{judge\}

\end{casebox}

\section{AI Tool Usage Disclaimer}\label{app:disclaimer}
AI-assisted tools are used solely for grammar correction, language polishing, and minor wording improvements in this manuscript. Their usage is strictly limited to linguistic and editorial assistance. All core ideas, methodological contributions, and experimental designs are developed entirely by the human authors. 

\newpage

\clearpage

\end{document}